\pdfoutput=1

\documentclass[11pt,table]{article}
\usepackage[preprint]{acl}
\usepackage{times}
\usepackage{latexsym}

\usepackage[T1]{fontenc}

\usepackage[utf8]{inputenc}

\usepackage{microtype}

\usepackage{inconsolata}

\usepackage{graphicx}
\usepackage{url}
\usepackage{enumitem}
\usepackage{setspace}
\usepackage{amsmath}
\usepackage{multirow}
\usepackage{multicol}
\usepackage{tabularx}
\usepackage{xcolor}
\usepackage{pifont}
\usepackage{array}
\usepackage{booktabs}
\usepackage{amssymb}
\usepackage{caption} 
\usepackage{subcaption} 
\usepackage{float} 
\usepackage{adjustbox} 
\usepackage[many]{tcolorbox}
\makeatletter
\newcommand{\rmnum}[1]{\romannumeral #1}
\newcommand{\Rmnum}[1]{\expandafter\@slowromancap\romannumeral #1@}
\makeatother
\definecolor{darkgreen}{rgb}{0.0, 0.5, 0.0}
\definecolor{lightgray}{gray}{0.9}
\definecolor{tabgray}{gray}{0.95}
\definecolor{lightorange}{rgb}{1.0,0.96,0.92}
\definecolor{FloralWhite}{rgb}{0.90,0.95,1.0}
\newcommand{\cmark}{\textcolor{darkgreen}{\ding{51}}} 
\newcommand{\xmark}{\textcolor{red}{\ding{55}}} 

\title{Pixel-Level Reasoning Segmentation via Multi-turn Conversations}

\author{Dexian Cai$^{1,\textsuperscript{*}}$, Xiaocui Yang$^{1,\textsuperscript{*}}$, Yongkang Liu$^1$, \textbf{Daling Wang}$^{1,\dag}$\\ 
\textbf{Shi Feng}$^{1}$, \textbf{Yifei Zhang}$^1$, \textbf{Soujanya Poria}$^2$\\
$^1$School of Computer Science and Engineering, Northeastern University, Shenyang, China \\
$^2$Singapore University of Technology and Design, Singapore \\
{\normalsize \texttt{2301840@stu.neu.edu.cn}}, ~~
{\normalsize \texttt{yangxiaocui@cse.neu.edu.cn}}, ~~
{\normalsize \texttt{misonsky@163.com}} \\
{\normalsize \texttt{\{wangdaling,fengshi,zhangyifei\}@cse.neu.edu.cn}, ~~
\normalsize \texttt{sporia@sutd.edu.sg}}
}

\begin{document}
\maketitle
\begin{abstract}
Existing visual perception systems focus on region-level segmentation in single-turn dialogues, relying on complex and explicit query instructions. Such systems cannot reason at the pixel level and comprehend dynamic user intent that changes over interaction. 
Our work tackles this issue by introducing a novel task, Pixel-level Reasoning Segmentation (Pixel-level RS) based on multi-turn conversations, tracking evolving user intent via multi-turn interactions for fine-grained segmentation. To establish a benchmark for this novel task, we build a \textbf{P}ixel-level \textbf{R}eason\textbf{I}ng \textbf{S}egmentation Dataset Based on Multi-\textbf{T}urn Conversations (PRIST), comprising 24k utterances from 8.3k multi-turn conversational scenarios with segmentation targets. Building on PRIST, we further propose MIRAS, a \textbf{M}ulti-turn \textbf{I}nteractive \textbf{R}e\textbf{A}soning \textbf{S}egmentation framework, integrates pixel-level segmentation with robust multi-turn conversation understanding, generating pixel-grounded explanations aligned with user intent.
The PRIST dataset and MIRSA framework fill the gap in pixel-level reasoning segmentation. Experimental results on the PRIST dataset demonstrate that our method outperforms current segmentation-specific baselines in terms of segmentation and LLM-based reasoning metrics. 
The code and data are available at: \url{https://github.com/ccccai239/PixelRIST}.
\end{abstract}
\def\thefootnote{\dag}\footnotetext{Corresponding author.} 
\def\thefootnote{\textsuperscript{*}}\footnotetext{Equal contribution.} 
\section{Introduction}
\begin{figure}[h!]
    \vspace{-3pt}
    \centering
    \captionsetup{skip=1.5pt}
    \includegraphics[width=1\linewidth]{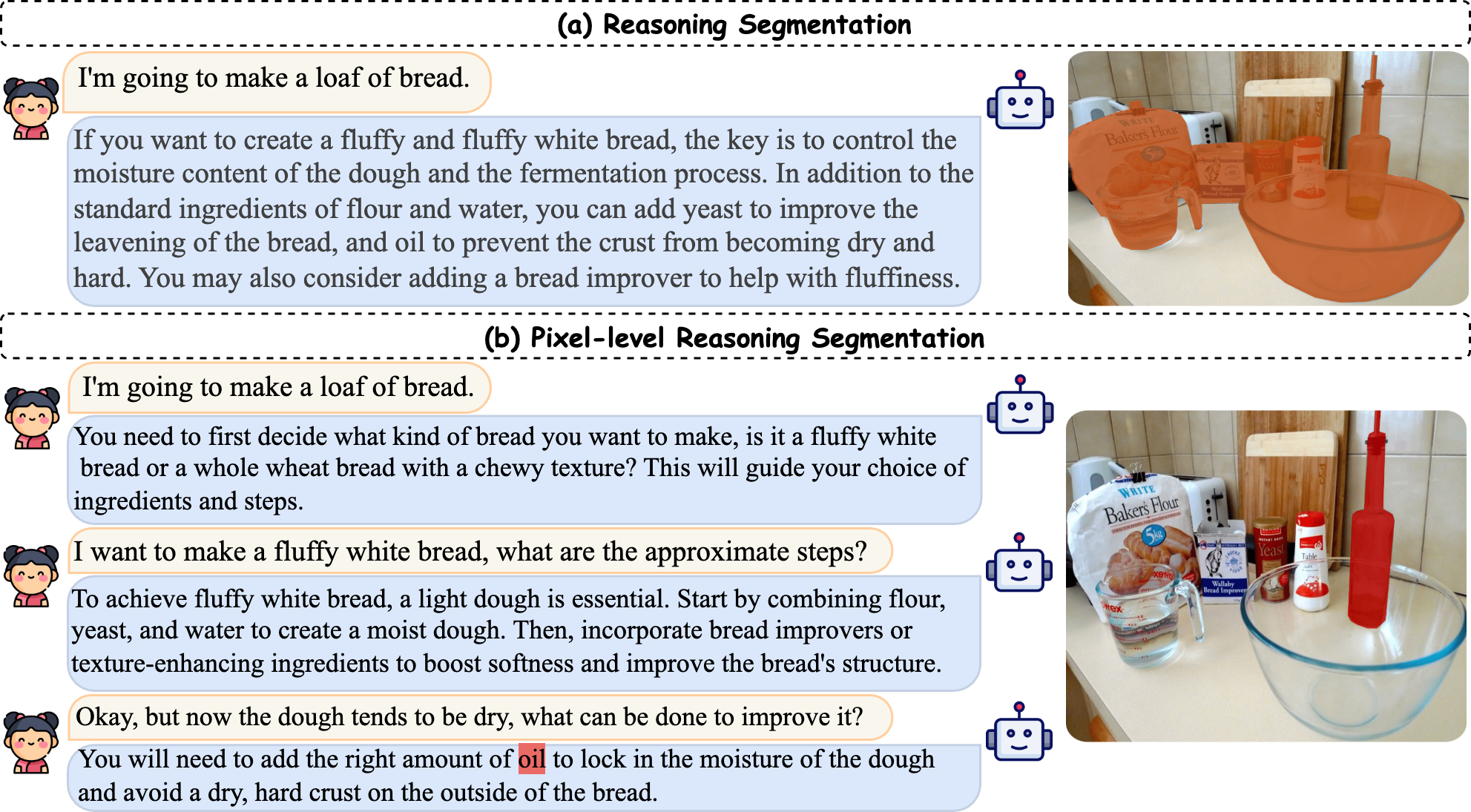}
    \caption{\textbf{RS vs. Pixel-level RS.} Pixel-level RS refines intent understanding and segmentation (\text{e.g.,} \textit{\textcolor{red}{"oil bottle"}}) through multi-turn interactions, while RS produces rough segmentation (\text{e.g.,} \textit{\textcolor{orange}{"all ingredients"}}) and handles implicit single-turn queries poorly.}
    \label{fig:intro}
    \vspace{-15pt}
\end{figure}

Existing general multimodal large language models (MLLMs) \cite{bai2023qwenvlversatilevisionlanguagemodel,zhu2023minigpt,liu2024visual} exhibit exceptional visual perception, enabling both image segmentation and textual reasoning, while they primarily rely on explicit human instructions for region-level grounding. 
Although some segmentation-specific works have explored grounded reasoning responses \cite{peng2023kosmos,you2023ferret,pi2023detgpt,zhang2023gpt4roi}, they depend on user-provided regions to trigger reasoning. These perception systems still cannot actively comprehend user's nuanced intent in real-world scenarios. To alleviate this problem, \citet{lai2023lisa} proposes the reasoning segmentation task that aims to achieve segmentation based on a implicit reasoning query. 
Recent studies \cite{ren2024pixellm,xia2024gsva,yuan2024osprey} have extended this region-level task to encompass multi-object segmentation scenarios to advance development. 
However, these methods have two limitations: 1) They rely on single-turn ambiguous queries and cannot fully understand users’ evolving intent. 2) They 
lack pixel-level segmentation and only achieve region-level segmentation through one-step explanations (e.g., segment roughly \textit{all ingridients} in Figure \ref{fig:intro}(a)). In contrast, multi-turn interactions can progressively clarify vague and generalized instructions such as \textit{"make a bread"}. 
As illustrated in Figure \ref{fig:intro}(b), the system through multi-turn interactions first guide to clarify the user's desired type of bread, providing targeted responses, and ultimately focuses on the user's specific needs, achieving pixel-level segmentation in final.

\setlength{\parskip}{0pt}
To address these challenges, we propose a novel task, Pixel-level Reasoning Segmentation (Pixel-level RS) based on multi-turn conversations, that refines both reasoning and segmentation through multi-turn interactions, requiring the system to understand the evolving user intent and generating pixel-level explanations, and segmentation masks.
Given the lack of benchmarks for pixel-level segmentation based on multi-turn reasoning, we build a \textbf{P}ixel-level \textbf{R}eason\textbf{I}ng \textbf{S}egmentation Dataset Based on Multi-\textbf{T}urn Conversation (\textbf{PRIST}), consisting of 24k utterances, 8.3k multi-turn conversational scenarios with specific segmentated targets, which provides a valuable resource for advancing Pixel-level RS research. PRIST focuses on pixel-level segmentation tasks while introducing new challenges in multi-turn reasoning and evolving intent comprehension. 
We design a progressive three-step dialogue automatic generation pipeline based on a reasoning tree to iteratively guide and generate dialogue content, inspired by Tree-of-Thought (ToT) \cite{yao2024tree}. By integrating a multi-step reasoning chain with a tree structure, this approach facilitates deeper and broader reasoning in pixel-level segmentation training.

\setlength{\parskip}{0pt}
To further advance this novel task, we propose a \textbf{M}ulti-turn \textbf{I}nteractive \textbf{R}e\textbf{A}soning \textbf{S}egmentation framework, \textbf{MIRAS}, that enables pixel-level segmentation through progressive reasoning.
MIRAS incorporates a dual-vision encoder that fuses multi-scale features to capture detailed visual information. To improve segmentation performance, we introduce a semantic region alignment strategy to inject semantic information into the mask decoder. 
Additionally, the framework supports multi-turn interactions to iteratively clarify user intent and ambiguous regions.
Given the inherent subjectivity in reasoning tasks, manual assessments can be influenced by personal preferences. To ensure fairness, we develop comprehensive evaluation metrics leveraging Large Language Models (LLMs) to assess multi-turn reasoning segmentation across coherence, consistency, and accuracy dimensions.
Our contributions can be summarized as follows:
\setlength{\parskip}{0pt}\begin{itemize}[topsep=0pt, partopsep=0pt, itemsep=0pt, parsep=0pt,left=0pt]
\item We propose a novel task, Pixel-level Reasoning Segmentation, that aims to achieve fine-grained segmentation through multi-turn conversations. Then, we build the PRIST dataset, including 8.3k high-quality multi-turn conversational scenarios and pixel-level segmentation targets. 
\item We develop a multi-turn reasoning segmentation framework, MIRAS, that facilitates pixel-level intentional understanding and segmentation through multi-turn interactions. 
\item Comprehensive experimental results of our proposed method on different metrics demonstrate both the utility of the PRIST dataset and the effectiveness of the model.
\end{itemize}
\begin{figure*}[htbp]
    \centering
    \captionsetup{skip=1.5pt}
    \includegraphics[width=1\linewidth]{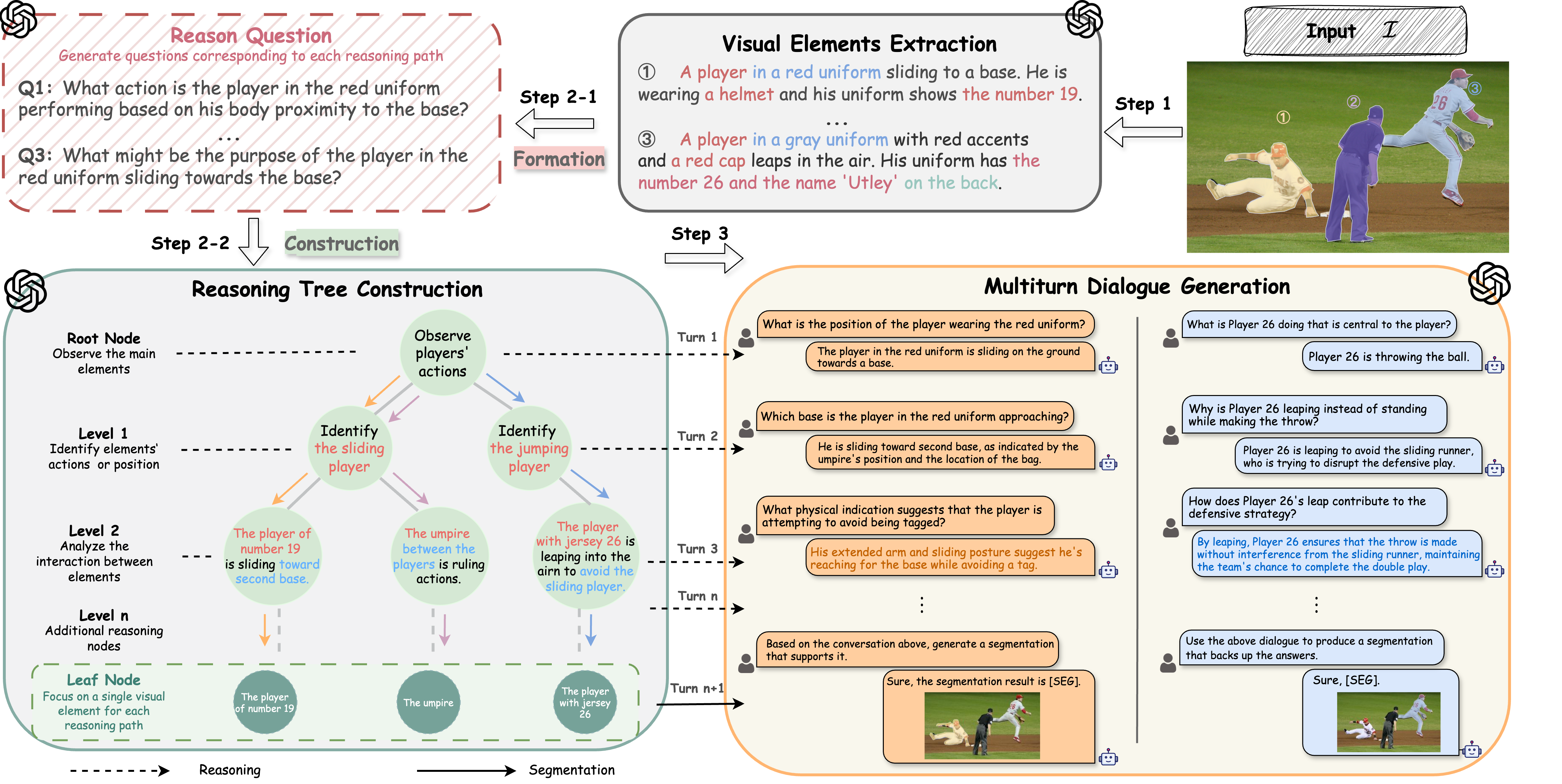}
    \caption{\textbf{The Generation Pipeline of PRIST Dataset.} \rmnum{1}) \textit{Step 1} extracts visible elements from images, establishing a semantic foundation for subsequent steps. \rmnum{2}) \textit{Step 2-1} generates complex reasoning questions from these elements, while \textit{Step 2-2} iteratively refines the questions through a reasoning tree, ensuring rigorous reasoning. \rmnum{3}) \textit{Step 3} organizes the nodes in reasoning tree into a multi-turn dialogue format.}
    \label{fig:pipeline}
    \vspace{-15pt}
\end{figure*}
\vspace{-5pt}
\section{Related Work}
\vspace{-5pt}
\subsection{Datasets for Reasoning Segmentation}
\vspace{-5pt}
Current reasoning segmentation datasets \cite{lai2023lisa,rasheed2024glamm,yuan2024osprey} mainly focus on region-level segmentation and single-step reasoning, which fail to meet the comprehensive requirements of the pixel-level RS task.
Traditional region-level segmentation datasets \cite{yu2016modeling,krishna2017visual} rely on simple and explicit instructions, lacking complex user intents understanding. To bridge this gap, ReasonSeg \cite{lai2023lisa} is the first to propose a segmentation dataset based on complex queries, which is a small scale and does not support multi-turn interactions. While recent datasets \cite{yuan2024osprey,rasheed2024glamm,ren2024pixellm} have expanded in size and diversity, they remain focused on multi-object segmentation and provide limitations for multi-step reasoning and fine-grained grounding. In contrast, our PRIST dataset utilizes conversation to simulate a human-like multi-step reasoning process, which innovatively combines multi-turn reasoning with pixel-level segmentation. 
Detailed dataset comparisons are shown in Table \ref{tab:compare_benchmark} in Appendix \ref{dataappendix}.
\vspace{-5pt}
\subsection{MLLMs for Region-level Segmentation} 
\vspace{-5pt}
MLLMs have advanced vision-language perception tasks, with recent works 
\cite{peng2023kosmos,you2023ferret,zhang2023gpt4roi} focusing on image-level visual dialogue. Some \cite{chen2023shikra,peng2023kosmos,pi2023detgpt} achieve region-level understanding by incorporating positional information and boundary boxes, mainly relying on LLMs for region interpretation. Several models \cite{lai2023lisa,ren2024pixellm,rasheed2024glamm,zhang2024omg} integrate segmentation-specific modules with LLMs for end-to-end training, enabling a more comprehensive understanding of regions. While these methods address pixel-level grounding, they still face limitations in complex reasoning. More comparison between models in Appendix \ref{sec:compare}. Our proposed MIRAS overcomes these challenges by enhancing segmentation accuracy through interactive reasoning, progressively refining the boundaries. 
\vspace{-2pt}
\section{PRIST}
\vspace{-5pt}
The pixel-level RS task takes an image $\mathcal{I}$ and a multi-turn dialogue $\mathcal{D}$ as input, then simultaneously generates a target segmentation mask $\mathcal{M}$ along with a textual reasoning chain $\{a_1,a_2,\dots,a_N\}$ that captures the complete dialogue history, where $a_i$ denotes the system's response in the $i$-th turn. It is defined as follows:

\begin{small}
\begin{equation}
\begin{aligned}
(\mathcal{M},\{a_1,a_2,\dots,a_N\})&= \mathbf{Model}(\mathcal{I},\mathcal{D}).\\
\end{aligned} 
\end{equation}
\end{small}%
Furthermore, we construct the pixel-level reasoning segmentation dataset based on the multi-turn conversation (PRIST) using a three-step progressive annotation pipeline, capturing fine-grained details through context-aware multi-turn dialogue. 
\vspace{-5pt}
\subsection{Data Preparation} 
\label{Data Preparation}
\vspace{-5pt}
Given the focus on pixel-level segmentation, we select TextCaps\footnote{TextCaps contains a total of 28k images.} \cite{sidorov2020textcapsdatasetimagecaptioning} as the image source due to its detailed visual information. To ensure a diverse range of scenes, we randomly select 280 images from each of the 10 major categories, resulting in a total of 2.8k images.
\vspace{-5pt}
\subsection{Generation Pipeline} \label{GenPipeline}
\vspace{-5pt}
We propose a three-step progressive automated annotation pipeline to create the PRIST Dataset, as illustrated in Figure \ref{fig:pipeline}. Appendix \ref{sec:appendixA.2} details the pipeline's prompts and output formats.
\vspace{-5pt}
\subsubsection{Visual Elements Extraction (Step-1)}
We first extract $N$ visible objects $\mathcal{O} = \{o_i\}_{i=1}^{N}$ 
, from the input image $\mathcal{I}$. Each object $o_i$ represents a distinct target for generating a dialogue. Specifically, we automatically identify visible elements with detailed attributes (e.g., color, position) to each object using GPT-4o \cite{achiam2023gpt}, along with corresponding textual descriptions (see Figure \ref{fig:sub1}). This step ensures that visual and semantic details are fully represented.
\vspace{-5pt}
\subsubsection{Reasoning Process Construction (Step-2)}
Pixel-level RS addresses complex scenarios requiring multi-turn reasoning with implicit user instructions. To simulate such scenarios, we implement a "question-first, problem-solving" strategy as shown in Figure \ref{fig:pipeline}, where the reasoning process is refined after first forming the reasoning problem and then constructing the reasoning tree. To align with multi-turn reasoning, we propose a hierarchical reasoning tree that recursively decomposes complex questions into smaller subquestions, progressively focusing on segmentation targets. Each reasoning tree path connects related elements to build a logical chain for pixel-level segmentation.

\vspace{-5pt}
\paragraph{Reasoning Question Formation (Step 2-1)}
This step expands a complex reasoning question $\mathcal{Q}_i$ for each target $o_i$, serving as the overall origin for next question decomposition and the theme for multi-turn dialogues in Step-3. To balance processing efficiency, we randomly select $K$ objects ($2 \leq K \leq \min(N, 4)$) from the element set $\mathcal{O}$ in Step-1 as targets, as fewer may missing essential interactions while more escalate complexity.
    
\vspace{-5pt}
\begin{small}
\begin{equation}
\begin{aligned}
\mathcal{Q}_i &=\mathbf{Formation}(\mathcal{I},o_i), i =\{1,..., K\}. \\
\end{aligned} 
\end{equation}
\end{small}%

\vspace{-5pt}
\paragraph{Reasoning Tree Construction (Step 2-2)}
The construction process (see Figure \ref{fig:sub2}) with GPT-4o initiates by establishing the elements $o_i$ as leaf nodes, allowing their corresponding reasoning problems $\mathcal{Q}_i$ to develop a distinct path $\mathcal{P}_i$ within the reasoning tree $\mathcal{T}$. Through iterative decomposition, each $\mathcal{Q}_i$ evolves into a sequence of progressive QA pairs $(q_n, a_n)$, with the tree's depth directly corresponding to the granularity of subquestions.
This hierarchical expansion refines the problem-solving framework and progressively narrows pixel-level target localization.
The resulting reasoning tree $\mathcal{T}$ explicitly captures the logical progression of complex questions, providing a structural foundation for the multi-turn dialogues in Step-3.

\vspace{-5pt}
\begin{small}
\begin{equation}
\begin{aligned}
\mathcal{P}_{i} &=  \mathbf{Construction}(\mathcal{Q}_i,o_i), \\
 &= \{(q_1,a_1), \dotsm,(q_n,a_n)\}, \\
 \mathcal{T} &= \{\mathcal{P}_{\text{1}},\mathcal{P}_{\text{2}},\dotsm,\mathcal{P}_{i}\}^{K}_{i=1},
\end{aligned} 
\end{equation}
\end{small}%
where $n$ is the depth of the reasoning path.
To manage computational resources and logical flexibility, we impose a constraint limiting each reasoning tree layer to a maximum of three child nodes.
\vspace{-5pt}
\subsubsection{Multi-turn Dialogue Generation (Step-3)}
We further build multi-turn dialogue $\mathcal{D}_i$ based on the hierarchical reasoning tree from Step-2.
Specifically, we integrate all nodes in each reasoning path $\mathcal{P}_i$, where each node represents a QA pair, to form the progressive multi-turn dialogue $\mathcal{D}_i$. Thus, each image can form $K$ conversations, $\{\mathcal{D}_1,\dotsm,\mathcal{D}_i\}^{K}_{i=1}$. To ensure responses fully integrate contextual information, prompts are designed to incorporate key elements and relationships from Step-1 (see Figure \ref{fig:sub3}), expanding understanding of landmarks, historical context, and scene interactions. 
Furthermore, pixel-level RS is designed to guide the model in performing fine-grained segmentation, with the final query in each dialogue being a segmentation-related instruction (e.g., \textit{"Please segment the core objects according to the above dialogue"}).
\begin{figure*}[htbp]
    \centering
    \captionsetup{skip=1.5pt}
    \includegraphics[width=1\linewidth]{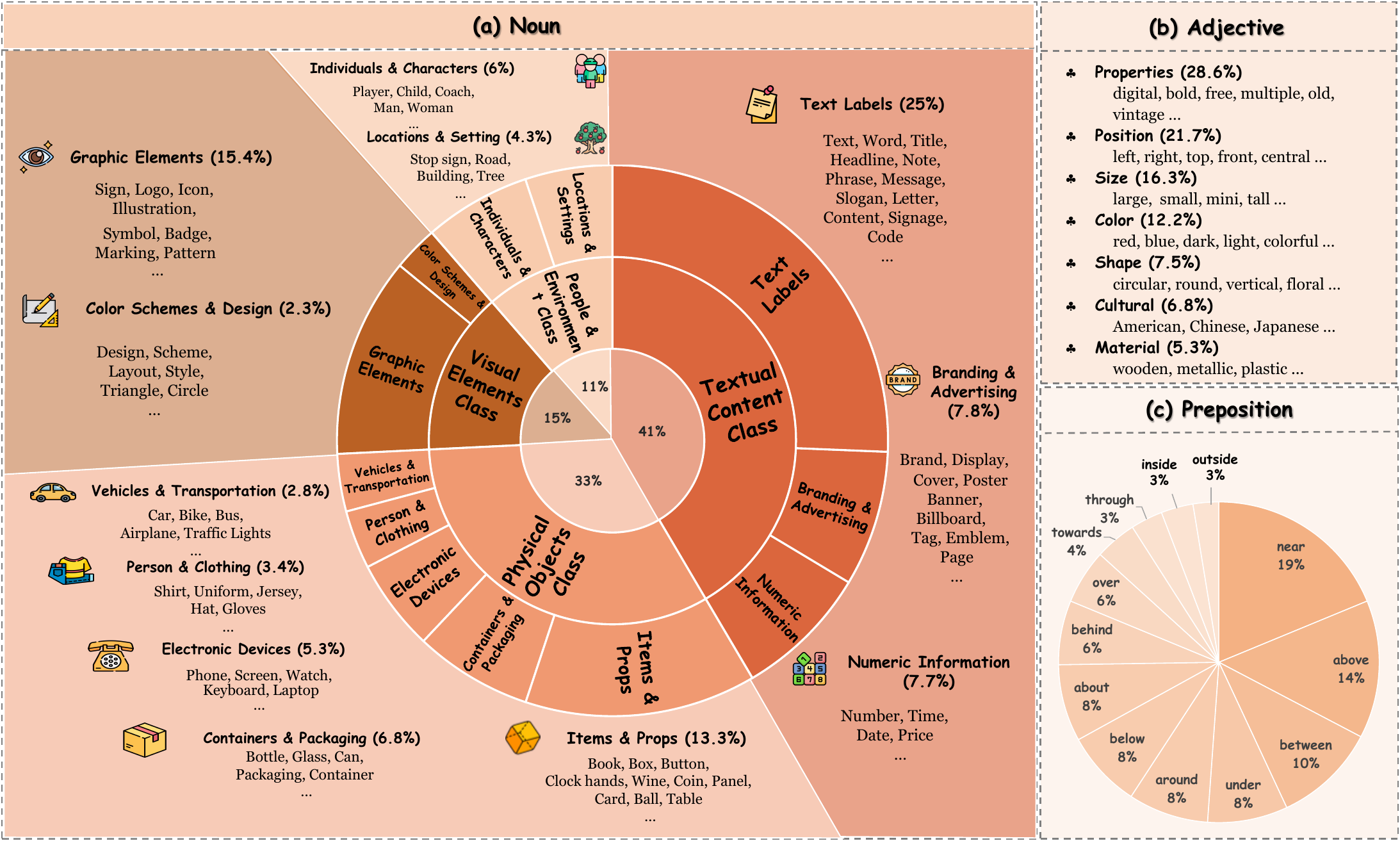}
    \caption{\textbf{The focus distribution of PRIST.} We analyze focus objects across 3 dimensions: noun, adjective and preposition, which capture fine granularity, diversity, and close spatial relationships between objects.}
    \label{fig:focus_stat}
    \vspace{-15pt}
\end{figure*}
\begin{table}[h!]
    \vspace{-8pt}
    \centering
    \small
    \captionsetup{skip=2pt}
    \setlength{\tabcolsep}{8pt} 
    \begin{tabularx}{0.43\textwidth}{lccccc}
    \toprule
     & \textbf{G1} & \textbf{G2} & \textbf{G3} & \textbf{G4} & \textbf{G5}\\ \midrule
    IoU &  0.82 & 0.88 & 0.91 & 0.86 & 0.81\\
    Kappa & 0.79 & 0.82 & 0.80 & 0.78 & 0.75 \\
    \bottomrule
    \end{tabularx}
    \caption{Consistency Analysis Results for Mask Annotators. G$x$ denotes the $x$-th expert group.}
    \label{tab:annotation}
    \vspace{-15pt}
\end{table}
\vspace{-3pt}
\subsection{Quality Assurance}
\vspace{-3pt}
We employ manual annotation to generate the true segmentation masks for each sample. Ten experts, selected through qualification tests, annotate masks and correct any commonsense errors in the dialogues. To ensure the quality of PRIST, we implement a double-check process. Specifically, experts are organized into five groups, with each group annotating the same set of samples. This set ensures that two annotators independently annotate every sample. More details are in Appendix \ref{sec:appendixA.1}.
The reliability of annotations across the five groups is assessed using IoU \cite{girshick2014rich}, which measures the overlap between two annotators, and Cohen’s Kappa \cite{cohen1960coefficient}, which quantifies their consistency. As shown in Table \ref{tab:annotation}, all groups achieve a result of IoU > 0.80 and Kappa > 0.75, demonstrating high annotation quality and strong consistency across the groups.
\vspace{-3pt}
\subsection{Dataset Analysis} 
\vspace{-3pt}
\label{DataAnalysis}
As shown in Table \ref{tab:stat}, we provide detailed statistics of the PRIST dataset.
PRIST contains 24k high-quality utterances, and 8.3k multi-turn conversational scenarios, with each scenario focusing on a single, fine-grained object. The dataset is divided into three subsets: train / validation / test splits, containing 7,312 / 500 / 508 samples, respectively. 

\begin{table}[t]
    \centering
    \captionsetup{skip=2pt}
    \small
    \setlength{\tabcolsep}{10pt} 
    \begin{tabular}{@{}ll@{}}
        \toprule
        \textbf{Statistic}                  & \textbf{Number} \\ \midrule
        Total Images                        & 2,800            \\
        Total Samples                       & 8,320            \\ \midrule
        \textbf{Segmentation}                                    \\
        - Focus Classes                     & 12               \\
        - Granularity (Coarse: Med.: Fine) & 22\%: 25\%: 53\%   \\ 
        \textbf{Multi-turn Dialogue}                             \\
        - Number of Utterances                & 24k               \\
        - Avg. / Max. Turns                   & 4 / 7             \\
        - Avg. / Max. Dialogue Length         & 477.6 / 518       \\ \bottomrule
    \end{tabular}
    \caption{\textbf{PRIST Statistics.} According to COCO’s image standard (640 vs. 1024), mask granularity is categorized as "Fine" (< $(s \times 32)^2$ px), "Med." ($(s \times 32)^2$ to $(s\times96)^2$ px), and "Coarse" (> $(s \times 96)^2$ px).} 
    \label{tab:stat}
    \vspace{-15pt}
\end{table}

To quantify pixel-level segmentation, we adopt the COCO mask size standard\footnote{\url{https://cocodataset.org/\#detection-eval}} for measuring target scales, aligning with existing datasets \cite{chen2015microsoft,caesar2018coco} and providing a quantitative basis for fine-grained segment evaluation. Statistics in Table \ref{tab:stat}, fine-grained targets (the scaling factor $s$ is 1.6) account for 53\% of PRIST, surpassing the 41\% of small targets in COCO \cite{chen2015microsoft}. With a minimum mask area of 304px and a standardized image size of 1024 × 1024, PRIST meets the fine-grained requirements of pixel-level segmentation.
We emphasizes exhibiting high diversity in categories and descriptions to enhance expressiveness. Illustrated the focus distribution of PRIST in Figure \ref{fig:focus_stat}, the categories include four types: \textit{Textual Content}, \textit{Physical Objects}, \textit{Visual Elements}, and \textit{People \& Environment}, which are further refined into 12 subcategories that cover objects from coarse- to fine-grained levels. At the descriptive level, a combination of \textit{Noun}, \textit{Adjective}, and \textit{Preposition} is employed: nouns provide basic category information (e.g., "tree"), adjectives enrich focus details (e.g., "worn-out chair"), and prepositions describe spatial relationships (e.g., "a book under the table"). 
PRIST delivers rich semantic-spatial annotations, establishing a benchmark resource for Pixel-level RS advancement.
\vspace{-3pt}
\section{MIRAS}
\vspace{-3pt}
To further research the novel pixel-level RS task, we propose MIRAS, a framework that refines user intent through multi-turn interactions to achieve pixel-grounded explanations and segmentation.
\vspace{-3pt}
\subsection{Architecture} 
\begin{figure*}
    \centering
    \captionsetup{skip=1pt}
    \includegraphics[width=1\linewidth]{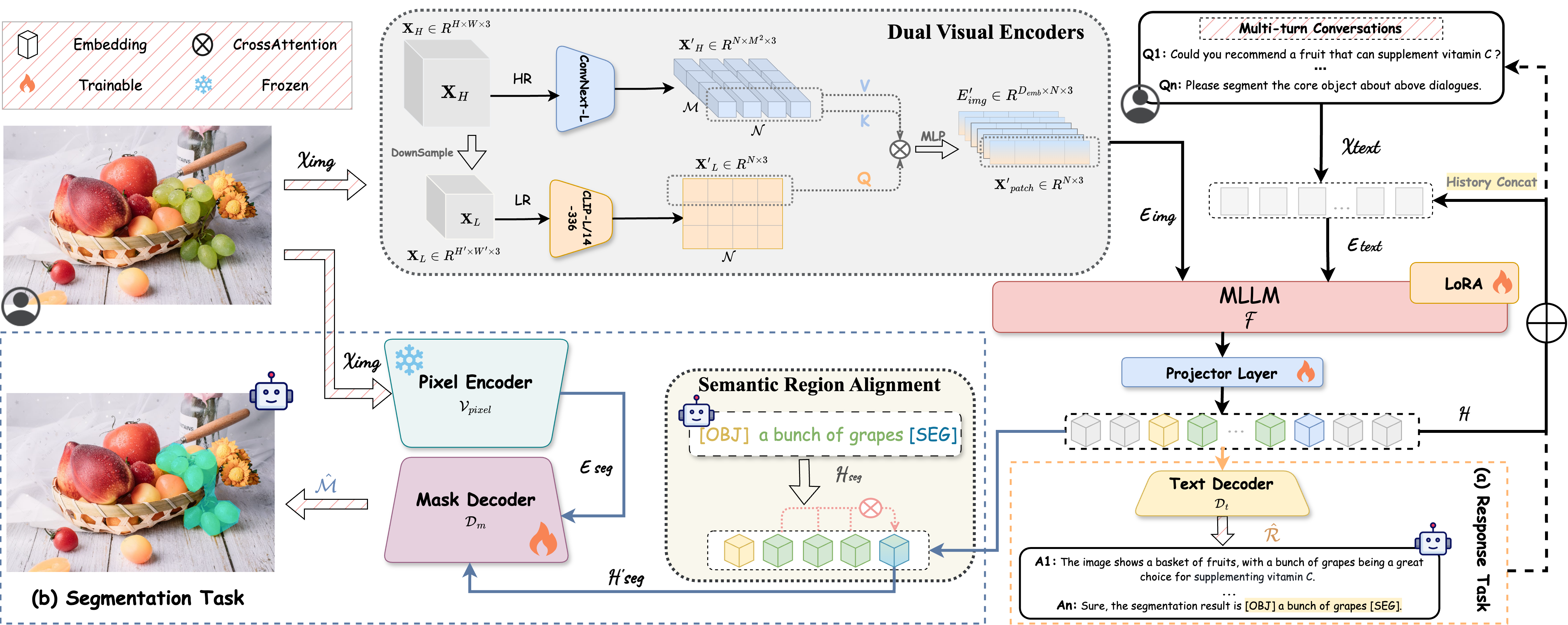}
    \caption{\textbf{Overview Architecture of MIRAS.} The model integrates MLLM and SAM modules by introducing a special token \texttt{[SEG]}. MIRAS can perform both (a) Multi-turn Response and (b) Segmentation tasks end-to-end.} 
    \label{fig:overview}
    \vspace{-15pt}
\end{figure*}

\vspace{-3pt}
\label{arch}
The architecture of MIRAS is illustrated in Figure \ref{fig:overview}, consisting of three core components: Visual Encoder, MLLM ($\mathcal{F}$, \cite{liu2024improved}), and Mask Decoder ($\mathcal{D}_m$, \cite{kirillov2023segany}). 
To seamlessly connect reasoning with segmentation, we introduce a special token \verb|[SEG]| as a placeholder for segmenting regions and enabling end-to-end processing. Two key modules are proposed for pixel-level RS. First, we integrate dual visual encoders \cite{li2024mgm} to extract enriched visual features. Then, the semantic region alignment strategy is designed to further refine the model's focus by incorporating target semantic information.

\paragraph{Dual Visual Encoders} 
We train the dual visual encoder using a high-resolution image ($\mathbf{X}_H$, 768 × 768 pixels) processed by ConvNext-L \cite{liu2022convnet2020s} paired with its low-resolution counterpart ($\mathbf{X}_L$, 336 × 336 pixels) processed by CLIP-L/14 \cite{radford2021learningtransferablevisualmodels}, downsampled from $\mathbf{X}_H$. Then, different resolutions are fused by a cross-attention module \cite{lin2022cat} to enhance visual detail capturing. Note that $\mathbf{X}_H$ equals $\mathbf{X}_\text{img}$.

\vspace{-8pt}
\begin{small}
\begin{equation}
\begin{aligned}
\mathbf{X}_H' &= \textbf{ConvNext}(\mathbf{X}_{H}), \quad \mathbf{X}_H' \in \mathbb{R}^{H \times W \times 3},\\
\mathbf{X}_L' &= \textbf{CLIP}(\mathbf{X}_L), \quad \mathbf{X}_L' \in \mathbb{R}^{H' \times W' \times 3},\\
\mathbf{E}'_{\text{img}} &=\text{CrossAtten} (Q=\mathbf{X}_{L}', K=\mathbf{X}_{H}', V=\mathbf{X}_H'), \\
\mathbf{E}_{\text{img}} &= \text{MLP}(\mathbf{E}'_{\text{img}})+\mathbf{E}'_{\text{img}}, \\
\end{aligned}
\end{equation}
\end{small}%

\vspace{-15pt}
\paragraph{Semantic Region Alignment}
We employ the SAM to obtain the pixel-level image features.

\vspace{-5pt}
\begin{small}
\begin{equation}
    \begin{aligned}
    \mathbf{E}_\text{seg} &= \mathcal{V}_{\text{pixel}}(\mathbf{X}_\text{img}) ,\quad \mathbf{E}_\text{seg} \in \mathbb{R}^{N\times256}.\\
    \end{aligned}
\end{equation}
\end{small}%
To provide clear segmentation intent, we design a novel segmentation prompt template \verb|[OBJ]{CLASS}[SEG]|, where \verb|{CLASS}| is the object description (e.g., \verb|[OBJ]| \verb|a bunch of grapes| \verb|[SEG]| in Figure \ref{fig:overview}). We further utilize special tokens \verb|[OBJ]| to extract relevant sub-sequence $\mathcal{H}_{\text{seg}}$ from $\mathcal{H}$ for segmentation and employ the cross-attention module to capture sufficient semantic information, which is crucial for efficient fine-grained segmentation. The method resolves potential mismatches in dimensions caused by varying lengths of \verb|{CLASS}|, denoted as $N^{sub}$.

\vspace{-5pt}
\begin{small}
\begin{equation}
    \begin{aligned}
    \mathcal{H}_{\text{seg}}' &= \text{CrossAtten}(Q, K ,V | \mathcal{H}_{\text{seg}}),  \\
     \mathcal{H}_{\text{seg}} &\in \mathbb{R}^{N^{sub} \times 256}, \mathcal{H}_{\text{seg}}' \in \mathbb{R}^{1\times 256},\\
    \end{aligned}
\end{equation}
\end{small}%

The mask decoder $\mathcal{D}_m$ combines region features from the pixel encoder with the hidden features of \verb|[SEG]| to produce the final mask. 

\vspace{-5pt}
\begin{small}
\begin{equation}
    \begin{aligned}
    \hat{\mathcal{M}} &= \mathcal{D}_m(\mathbf{E}_\text{seg},\mathcal{H}_{seg}'),\\
    \end{aligned}
\end{equation}
\end{small}
\begin{table*}[htbp]
\centering
\captionsetup{skip=1.5pt}
\small
\setlength{\tabcolsep}{4.25pt} 
\begin{tabularx}{\textwidth}{lcccc|ccccc}
\toprule
\multirow{2}{*}{\textbf{Model}} & \multirow{2}{*}{\textbf{CIoU}} & \multicolumn{3}{c}{\textbf{Pixel-wise}} & \multicolumn{5}{c}{\textbf{Response}} \\
\cmidrule(lr){3-5} \cmidrule(lr){6-10} 
 & & Prec. & Recall & F1 & BLEU-4 & Dist-1/2 & ROU\_L. & MET. & BERTS.\\
\midrule 
\rowcolor{tabgray}
\multicolumn{10}{c}{\emph{Zero-shot}} \\ \midrule
InternVL2-8B* \cite{chen2024far} & 8.26 & 7.59 & 11.55 & 9.16 & 1.20 & 7.5 / 41.0 & 18.44 & 23.98 & 84.78 \\
Qwen2-VL-7B* \cite{Qwen2VL} & 10.64 & 10.30 & 15.88 & 12.50 & 3.00 & 9.3 / 41.2 & 26.28 & 28.26 & 86.73 \\
LLaVA-v1.5-7B* \cite{liu2024visual} & 11.25 & 11.35 & 25.90 & 15.68 & 3.21 & 5.6 / 27.5 & 16.02 & 18.41 & 78.81 \\
LLaVA-v1.6-7B* \cite{liu2024improved} & 11.84 & 11.90 & 34.78 & 17.69 & 1.07 & 5.5 / 27.5 & 20.60 & 25.00 & 84.26 \\ 
GPT-4o* \cite{openai2024hello} & \underline{14.13} & 17.35 & 35.01 & 23.18 & 4.30
& 9.1 / 42.9 & 26.35 & 28.55 & 87.62 \\
\midrule
LISA \cite{lai2023lisa}& 10.45 & 15.33 & 43.07 & 15.09 & 1.97 & 6.2 / 28.4 & 18.21 & 26.59 & 85.67 \\
PixelLM \cite{ren2024pixellm}& 9.87 & 17.21 & 35.36 & 14.68 & 1.34 & 6.7 / 30.1 & 14.93 & 21.26 & 85.13 \\
OMG-LLaVA \cite{zhang2024omg}& 9.67 & 16.67 & 77.80 & 27.46 & 8.70 & 11.2 / 42.0 & 23.47 & 27.90 & 87.30 \\
\rowcolor{lightorange}
MIRAS (\textit{Stage-1}) & 13.12 & 15.64 & 45.11 & 23.22 & 4.17 & 6.9 / 28.4 & 25.94 & 28.18 & 87.54 \\ \midrule
\rowcolor{tabgray}
\multicolumn{10}{c}{\emph{Fine-tuning}} \\ \midrule
LISA \cite{lai2023lisa} & 11.23 & \textbf{26.23} & 29.22 & 27.64 & 7.81 & \underline{14.2 / 40.7} & 27.84 & 30.74 & 86.75 \\ 
PixelLM \cite{ren2024pixellm}& 10.32 & 20.95 & 18.84 & 11.71 & \underline{9.97} & 11.6 / 38.0 & \underline{30.63} & 32.99 & 87.80 \\ 
OMG-LLaVA \cite{zhang2024omg}& 13.84 & 21.54 & 49.31 & \underline{29.98} & \textbf{11.21} & 12.3 / 35.3 & 30.59 & \underline{39.18} & \textbf{88.76} \\ 
\rowcolor{lightorange}
MIRAS (\textit{Stage-2}) & \textbf{14.72} & \underline{24.22} & 40.61 & \textbf{30.34} & 8.51 & \textbf{15.7 / 49.2} & \textbf{30.82} & \textbf{40.06} & \underline{88.47} \\ 
\bottomrule
\end{tabularx}
\caption{\textbf{Results on Pixel-level RS.} * denotes the general MLLMs, others are 7B segmentation-specific MLLMs.
Bold text indicates the best result and underlined text indicates suboptimal results.
}
\label{tab:TreeMDS}
\vspace{-15pt}
\end{table*}
\vspace{-10pt}
\subsection{Training Process} 
\vspace{-3pt}
\label{training}
We employ a two-stage training process to achieve efficient pixel-level reasoning segmentation. In Stage-1, mask-text alignment pretraining based on various datasets is conducted, followed by instruction fine-tuning using the PRIST dataset in Stage-2 (more details in Appendix \ref{two-stage}). The objectives remain consistent across both stages: the text generation loss $\mathcal{L}_{\text{t}}$ and a linear combination of per-pixel BCE loss $\mathcal{L}_{bce}$ and DICE loss $\mathcal{L}_{dice}$ for segmentation. Only the mask decoder and projection layer are trainable to balance efficiency and performance while keeping the image encoder and MLLM frozen. The training loss is formulated as:

\vspace{-5pt}
\begin{small}
\begin{equation}
\begin{aligned}
\mathcal{L} = \lambda_{t}\mathcal{L}_{\text{t}}(\mathcal{R}, \hat{\mathcal{R}}) + \lambda_{bce}{BCE}(\mathcal{M}, \hat{\mathcal{M}}) \\ + \lambda_{dice}{DICE}( \mathcal{M}, \hat{\mathcal{M}})
\end{aligned}    
\end{equation}
\end{small}%
where $\lambda_{\text{t}}$, $\lambda_{\text{bce}}$ and $\lambda_{\text{dice}}$ values 1.0, 2.0 and 0.5 separately, following LISA \cite{lai2023lisa}.

\vspace{-5pt}
\section{Experiment}
\vspace{-5pt}
We conduct experiments to assess our model's adaptability to the novel pixel-level RS task and its general performance in the classical referring expression segmentation (RES) task \cite{kazemzadeh-etal-2014-referitgame,yu2016modeling}. 
More implementation details are in Appendix \ref{setups}. 
\vspace{-3pt}
\subsection{Evaluation Metrics} 
\vspace{-3pt}
\label{metrics}
We establish a benchmark for Pixel-level RS across three evaluation aspects: pixel-level segmentation, conversation response, and reasoning quality. For segmentation, we employ the CIoU metric \cite{zheng2020distance} and propose pixel-wise mask precision, recall, and F1 metrics. The precision measures the accuracy of segmentation while recall evaluates the coverage. 
For response, the metrics including BLEU-4 \cite{papineni2002bleu}, ROUGE-L \cite{kingma2014adam}, Dist-n \cite{li2015diversity}, and METEOR \cite{banerjee2005meteor}. Considering the subjectivity of reasoning, we introduce LLM as a scoring tool using four metrics: Progressiveness \textbf{(PR)}, Logical Coherence \textbf{(LC)}, Content Consistency \textbf{(CC)}, and Target Relevance \textbf{(TR)}, with higher scores reflecting better performance across aspects. Meanwhile, we employ GPT-4o as a judge to assess dialogue reasoning quality. The model wins when its score surpasses that of the human response, as reflected by the \textbf{Win Rate (\%)} metric. Evaluation criteria are detailed in Appendix \ref{sec:appendixB.2}.

\vspace{-3pt}
\subsection{Baselines}
\vspace{-3pt}
We compare with three types of baselines: \textbf{1) General MLLMs.} We take advanced close- and open-source MLLMs to evaluate zero-shot on PRIST for pixel-level RS capability. \textbf{2) Segmentation-specific MLLMs.} LISA \cite{lai2023lisa}, PixelLM \cite{ren2024pixellm} and OMG-LLaVA \cite{zhang2024omg} are evaluated under zero-shot and fine-tuning settings to show PRIST's task-specific enhancement. \textbf{3) Segmentation-specific Models.} We compare three semantic segmentation models (\text{e.g.,} LVIT \cite{yang2022lavt}) on the RES task with MIRAS to verify its advantage of the architecture.
\vspace{-3pt}
\subsection{Experimental Results and Analysis}
As shown in Table \ref{tab:TreeMDS}, we comprehensively compare model performance across two critical dimensions: pixel-level segmentation and conversation response. Our proposed MIRAS demonstrates robust performance in pixel-level RS task, surpassing baselines and the closed-source GPT-4o in both segmentation and response metrics, establishing a new benchmark for the PRIST dataset. Notably, the performance improvement from PRIST fine-tuning exhibits universal applicability across different architectures. Table \ref{tab:winrate} further validates the enhanced reasoning capabilities of our method, with evaluation results approaching human expert proficiency.
\vspace{-15pt}
\subsubsection{Pixel-level Segmentation}
As illustrated in the left part of Table \ref{tab:TreeMDS}, GPT-4o achieves CIoU 14.13 and precision 17.35, surpassing open-source models. While it surpasses stage-1 MIRAS, it remains below fine-tuned MIRAS (\textit{stage-2}), indicating our framework can achieve closed-source model competency with efficient resource utilization. To enhance pixel-level performance, all segmentation-specific MLLMs are fine-tuned on the PRIST dataset under a consistent setting. Fine-tuning results in an increase in terms of CIoU and Precision, with OMG-LLaVA's CIoU $\textcolor{red}{\uparrow} 43\%$, Precision $\textcolor{red}{\uparrow} 29\%$, and LISA's 
Precision from 15.33 to 26.23 ($\textcolor{red}{\uparrow} 71\%$), respectively. MIRAS (\textit{stage-2}) establishes new benchmarks with precision 24.22, F1 30.34, and CIoU 14.72, demonstrating exceptional boundary delineation capabilities. Notably, this performance enhancement accompanies a precision-recall trade-off, i.e., recall decreases (LISA $\textcolor{blue}{\downarrow} 32\%$, OMG-LLaVA $\textcolor{blue}{\downarrow} 37\%$).

In zero-shot settings, all general MLLMs exhibit limited pixel-level segmentation, with GPT-4o slightly outperforming segmentation-specific MLLMs like PixelLM (Precision 17.35 vs. 17.21). This indicates that general MLLMs emphasize cross-task adaptability, while task-specific improvements rely on design-specific architecture (e.g., Mask Decoder). Additionally, the precision-recall trade-off observed in fine-tuned models reflects a strategic prioritization of segmentation specificity over generalizability in fine-grained tasks, which avoids the overgeneralization issues encountered in zero-shot settings, aligning with the objectives of pixel-level RS. An optimization choice validated by case studies in Appendix \ref{casestudies}.
\vspace{-3pt}
\subsubsection{Conversation Response}
In the right of Table \ref{tab:TreeMDS}, GPT-4o demonstrates near-expert dialogue competence across metrics, approaching fine-tuned MLLMs, such as OMG-LLaVA, while outperforming open-source general MLLMs like Qwen2-VL-7B. MIRAS achieves the best performance in metrics such as Dist-1/2 (15.7/49.2), ROUGE\_L (30.82), and METEOR (40.06), validating its ability to generate high-quality textual responses. GPT-4o’s strong baseline performance underscores its inherent dialogic intelligence; however, domain adaptation remains essential for optimal performance. Most fine-tuned models show improvements in text metrics, highlighting PRIST's effectiveness in bridging visual-textual semantic gaps. This demonstrates the framework’s dual competence in simultaneously optimizing mask-text alignment and response coherence.
\vspace{-15pt}
\subsubsection{Reasoning Quality}
Table \ref{tab:winrate} presents the reasoning quality evaluation based on LLMs (human scores detailed in Appendix \ref{human}). Fine-tuning on the PRIST dataset led to improvements across all models, with an average Win Rate increase of approximately 10\%. MIRAS achieved the SOTA with a Win Rate of 42\% and the highest scores in all four reasoning metrics, closely approaching human expert levels. The overall improvement in four fine-tuned MLLMs' reasoning quality shows the substantial potential of PRIST dataset in enhancing reasoning capabilities, stemming from the extensive conceptual vocabulary it provides during fine-tuning.
\begin{table}[htbp]
\vspace{-8pt}
\centering
\captionsetup{skip=1pt}
\tiny
\setlength{\tabcolsep}{5.4pt} 
\begin{tabularx}{0.40\textwidth}{lccccc}
\toprule
\textbf{Model} & \textbf{PR} & \textbf{LC} & \textbf{CC} & \textbf{TR} & \textbf{Win Rate(\%)}\\
\midrule
Human & 4.03 & 4.04 & 4.26 & 4.28 & - \\ \midrule
LISA(ft) & 3.76 & 3.69 & 3.71 & 3.58 & 36 ($\textcolor{red}{\uparrow}11$) \\
PixelLM(ft) & 3.35 & 3.48 & 3.32 & 3.28 & 32 ($\textcolor{red}{\uparrow}13$) \\
OMG-LLaVA(ft) & 2.60 & 2.48 & 2.58 & 2.33 & 24 ($\textcolor{red}{\uparrow}8$) \\
\rowcolor{lightorange}
MIRAS(\textit{Stage-2}) & \textbf{3.90} & \textbf{3.76} & \textbf{3.83} & \textbf{3.69} & \textbf{42 ($\textcolor{red}{\uparrow}$11)} \\
\bottomrule
\end{tabularx}
\caption{Comparison of the reasoning quality of domain-specific MLLMs fine-tuned on PRIST.} 
\label{tab:winrate}
\vspace{-15pt}
\end{table}
\vspace{-3pt}
\subsection{Generalization Segmentation}
We compare segmentation-specific baselines on classical referring expression segmentation benchmarks to evaluate the generalizability of MIRAS. As detailed in Table \ref{tab:stage1_res}, MIRAS' base configuration (last row) outperforms segmentation-specific models and demonstrates competitiveness against other MLLMs, even surpassing the latest OMG-LLaVA. Two findings emerge in results: (1) While fine-tuned MIRAS (\textit{Stage-2}) shows a decline in general performance due to task-specific optimization, it retains advantages over Next-Chat and remains comparable to OMG-LLaVA. (2) The capacity of the base model determines the system's potential, evidenced by consistent improvements in MIRAS when evolving the foundational model from LLaVA-v1 (row 6) to LLaVA-v1.6 (row 9).

\begin{table}[htbp]
\vspace{-8pt}
\tiny
\captionsetup{skip=1.5pt}
\setlength{\tabcolsep}{2pt} 
\begin{tabularx}{0.49\textwidth}{lcccccccc}
\toprule
\multirow{2}{*}{\textbf{Model}} & \multicolumn{3}{c}{\textbf{refCOCO}} & \multicolumn{3}{c}{\textbf{refCOCO+}} & \multicolumn{2}{c}{\textbf{refCOCOg}} \\
\cmidrule(lr){2-4} \cmidrule(lr){5-7} \cmidrule(lr){8-9}
 & Val & TestA & TestB & Val & TestA & TestB & Val & Test\\
\midrule
GRIS \cite{wang2022cris} & 70.5 & 73.2 & 66.1 & 65.3 & 68.1 & 53.7 & 59.9 & 60.4\\
LAVT \cite{yang2022lavt} & 72.7 & 75.8 & 68.8 & 62.1 & 68.4 & 55.1 & 61.2 & 62.1\\
GRES \cite{liu2023gres} & 73.8 & 76.5 & 70.2 & 66.0 & 71.0 & 57.7 & 65.0 & 66.0\\ \midrule
LISA (v1) \cite{lai2023lisa} & 74.1 & 76.5 & 72.3 & 65.1 & 70.8 & 58.1 & 67.9 & 70.6\\
PixelLM (v1) \cite{ren2024pixellm} & 73.0 & 76.5 & 68.2 & 66.3 & 71.7 & 58.3 & 69.3 & 70.5\\
\rowcolor{lightorange}
MIRAS (\textit{Stage-1}) (v1) & 75.3 & 78.9 & 70.2 & 66.7 & 74.3 & 61.5 & \textbf{73.2} & 71.9\\
OMG-LLaVA \cite{zhang2024omg} & \underline{78.0} & \underline{80.3} & \textbf{74.1} & \underline{69.1} & 73.1 & \underline{63.0} & \underline{72.9} & \textbf{72.9}\\ 
\rowcolor{lightorange}
MIRAS (\textit{Stage-2}) (v1.6) & 76.9 & 79.8 & 72.8 & 68.8 & \underline{74.4} & 62.5 & 71.8 & 70.8\\
\rowcolor{lightorange}
MIRAS (\textit{Stage-1}) (v1.6) & \textbf{78.4} & \textbf{80.5} & \underline{73.4} & \textbf{72.1} & \textbf{74.8} & \textbf{63.4} & 72.6 & \underline{72.0}\\
\bottomrule
\end{tabularx}
\caption{\textbf{Results on the RES benchmark.} "v1/v1.6" indicates LLaVA version.}
\label{tab:stage1_res}
\vspace{-15pt}
\end{table}

These impressive results are mainly attributed to MIRAS's convolutional backbone (\text{i.e.,} ConvNeXt), which supports larger input images and enables semantic-assisted mask generation. This provides a solid foundation for achieving fine-grained segmentation in the next stage. However, this focus on task-specific patterns inherently introduces a trade-off, sacrificing some degree of generalization.
\vspace{-8pt}
\subsection{Ablation Study}
\vspace{-3pt}
To validate the effectiveness of the modules in MIRAS, we conduct the following ablation experiments on the general RES task, ensuring compatibility with the following framework by maintaining the base model as LLaVA-v1. \textbf{(1) Dual-visual Encoder} ~ Table \ref{tab:ablation} illustrates that the dual-visual encoder improves performance (\textit{Val} $\downarrow2.6\%$, \textit{Test} $\downarrow0.9\%$) by supporting a higher resolution, which enhances the density of visual features and the ability to capture finer details. \textbf{(2) Semantic Region Alignment} ~ The alignment strategy of injecting semantic information has achieved positive results, as shown in Table \ref{tab:ablation}. When applied to the half dataset, it decreases \textit{Val} by 1.4\% and \textit{Test} by 0.5\%. Reducing the application to the full leads to further decline (\textit{Val} $\downarrow$0.7\%, \textit{Test} $\downarrow$0.4\%), highlighting its effectiveness in enhancing segmentation.

\begin{table}[htbp]
\centering
\captionsetup{skip=1.5pt}
\small
\setlength{\tabcolsep}{3pt} 
\begin{tabularx}{0.48\textwidth}{lcc}
\toprule
\multirow{2}{*}{\textbf{Architecture}} & \multicolumn{2}{c}{\textbf{refCOCOg}}  \\
 \cmidrule(lr){2-3}
 & Val (U) & Test (U)\\
\midrule
MIRAS (v1) & \textbf{73.2} & \textbf{71.9} \\
~ w/o Dual-visual Encoder & 70.6 & 70.8 \\
~ w/o Semantic Region Alignment (50\%) & 71.8 & 71.4 \\
~ w/o Semantic Region Alignment & 71.1 ($\downarrow$) & 71.0 ($\downarrow$) \\
\bottomrule
\end{tabularx}
\caption{\textbf{Ablation.} The metric is CIoU. 50\% means half of the samples randomly added semantic information.} 
\label{tab:ablation}
\vspace{-15pt}
\end{table}
\vspace{-5pt}
\section{Conclusion}
\vspace{-3pt}
In this paper, we propose a novel task, Pixel-level Reasoning Segmentation, which focuses on fine-grained segmentation. To further advance, we construct a pixel-level reasoning segmentation dataset, PRIST, consisting of 24k utterances and 8.3k pixel-level segmentation targets, generated through a carefully designed three-stage progressive automatic annotation pipeline. Additionally, we present MIRAS, a framework designed for this task that combines segmentation with multi-turn interaction, along with LLM-based reasoning quality evaluation metrics. 
Comprehensive experiments on segmentation and reasoning demonstrate the effectiveness of the PRIST dataset and the superior performance of MIRAS, which advances research in pixel-level reasoning segmentation meaningfully.

\section*{Limitations}
Although our research has achieved certain advancements in the pixel-level RS task, some limitations remain. PRIST is designed only for single-target segmentation, making it difficult to adapt to more complex scenarios, such as those with empty targets (i.e., no objects requiring segmentation) or multiple targets (i.e., simultaneously involving multiple distinct objects). Further exploring reasoning trees to model relationships among image elements and constructing datasets for multi-object, multi-level segmentation hold research potential. Additionally, we utilize the SAM model to efficiently assist MLLM in integrating text reasoning to achieve pixel-level segmentation. However, their integration of separate visual encoding modules creates structural redundancy, reducing efficiency. Developing a streamlined and efficient model architecture is an important direction for future work.

\section*{Ethics Statement}

\section*{Acknowledgments}

\bibliography{custom}

\appendix
\section{Segmentation Task}
\subsection{Referring Expression Segmentation (RES)}
The referring expression segmentation (RES) task \cite{kazemzadeh-etal-2014-referitgame} involves receiving an image and a natural language expression referring to a specific object in the image (e.g., \textit{"Please segment the apple in the image."}) as input, and then outputting the segmentation mask of that object. As a classic task in the field of semantic segmentation, it intuitively reflects the model's ability in visual localization. Refcoco, Refcoco+, and Refcocog provide mature evaluation benchmarks for this task. To ensure fairness in the comparison, we choose to compare segmentation-specific baseline models and conduct ablation studies of the model architecture on the RES task. This is because all segmentation-specific baselines support this task and are trained on the aforementioned datasets.
\subsection{Comparisons of Segmentation-specific MLLMs} \label{sec:compare}
Table \ref{tab:compare} provides a comprehensive comparison of recent segmentation-specific MLLMs in terms of model architecture, pixel-level capabilities, and conversation capabilities. A few works \cite{zhang2023gpt4roi,lai2023lisa,ren2024pixellm,rasheed2024glamm} integrate specialized vision modules and LMMs, as indicated by the Region Enc. / Dec. The End-End Model \cite{pi2023detgpt,chen2023shikra,peng2023kosmos} distinction separates models that leverage LMMs for region-level understanding from those employing external modules.
\begin{table}[h!]
\centering
\tiny
\setlength{\tabcolsep}{0.6pt} 
\begin{tabularx}{0.48\textwidth}{l*{6}{c}}
\toprule
\textbf{Method} & \multicolumn{2}{c}{\textbf{Region}} & \textbf{Pixel-Level} & \textbf{Conversation} & \textbf{End-End} \\
 & Enc. / Dec. & & Seg. / Cap. & Multi-turn / Reason & & \\
\midrule
VisionLLM \cite{wang2024visionllm} & \xmark/\xmark & & \xmark/\xmark & \xmark/\xmark & \xmark \\
DetGPT \cite{pi2023detgpt} & \xmark/\xmark & & \xmark/\xmark & \cmark/\cmark & \cmark \\
Shikra \cite{chen2023shikra} & \xmark/\xmark & & \xmark/\cmark & \xmark/\xmark & \cmark \\
Kosmos-2 \cite{peng2023kosmos} & \xmark/\xmark & & \xmark/\cmark & \xmark/\xmark & \cmark \\
GPT4RoI \cite{zhang2023gpt4roi} & \cmark/\xmark & & \xmark/\cmark & \cmark/\xmark & \cmark \\
LISA \cite{lai2023lisa} & \xmark/\cmark & & \cmark/\xmark & \xmark/\cmark & \xmark \\
PixelLM \cite{ren2024pixellm} & \xmark/\cmark & & \cmark/\xmark & \xmark/\cmark & \cmark \\
GLaMM \cite{rasheed2024glamm} & \cmark/\cmark & & \cmark/\cmark & \xmark/\cmark & \cmark \\
OMG-LLaVA \cite{zhang2024omg} & \xmark/\cmark & & \cmark/\cmark & \xmark/\cmark & \cmark \\
\rowcolor{lightgray}
MIRAS (Ours) & \xmark/\cmark & & \cmark/\cmark & \cmark/\cmark & \cmark \\
\bottomrule
\end{tabularx}
\caption{Comparison of recent Segmentation-specific MLLMs.}
\label{tab:compare}
\end{table}

\section{More Details about PRIST}
\subsection{Data Annotation} \label{sec:appendixA.1}
We recruit experts from the Computer Science department as annotators for our project, as they are familiar with the task requirements and objectives. Before starting annotation, all annotators underwent training and a small subset of data was pre-annotated, which was only considered a pass if the accuracy rate was at least 80\%. The subsequent annotation process was carried out once the pre-annotation results met the required standards. Additionally, We used the open-source tool Labelme\footnote{\url{https://github.com/wkentaro/labelme}}, which supports precisely outlining objects in arbitrary shapes and extracting masks, meeting the project’s needs for accuracy.

\noindent\textbf{Annotation process:} To avoid hallucinations and errors in the model-generated dialogues, we enforced strict quality control on the texts generated by GPT-4o. A total of 10 annotators were recruited, with a two-layer pyramid structure employed to reduce subjective bias and enhance quality. First, six annotators were divided into three pairs, each independently annotating different data subsets. Next, an additional annotator per group resolved inconsistencies and checked the quality of consistent labels. Finally, an experienced annotator consolidated and reviewed all data to ensure high quality and consistency. Each annotator’s main tasks included generating masks for target objects in images and correcting any commonsense errors in the dialogue content, such as mismatches in time or numbers between text and images.
\begin{table*}
\centering
\small
\setlength{\tabcolsep}{3.2pt} 
\begin{tabularx}{0.995\textwidth}{l*{10}{c}}
\toprule
\multirow{2}{*}{\textbf{Benchmark}} & \multirow{2}{*}{\textbf{\#Img.}} & \multirow{2}{*}{\textbf{\#Reg.}} & \multirow{2}{*}{\textbf{\#Samp.}} & \multirow{2}{*}{\textbf{Caption}} & \multicolumn{2}{c}{\textbf{Conversation}} & \multicolumn{2}{c}{\textbf{Segmentation}} & \multirow{2}{*}{\textbf{Reasoning}} & \multirow{2}{*}{\textbf{\#Step}}\\
 & & & & & Single. / Multi. & & Region. / Pixel. & & & \\
\midrule
\rowcolor{lightgray}
\multicolumn{11}{c}{\emph{Segmentation Benchmark}} \\
RefCOCO \cite{yu2016modeling} & 20K & 142K & 142K & \cmark & \cmark/\xmark & &\cmark/\xmark & & \xmark & -\\
VG \cite{krishna2017visual} & 82.4K & 3.8M & 3.8M & \cmark & \cmark/\xmark & &\cmark/\xmark & & \xmark & -\\
ReasonSeg \cite{lai2023lisa} & 1.2K & 1.2K & 1.2K & \cmark & \cmark/\xmark & &\cmark/\xmark & & \cmark & - \\
Osprey \cite{yuan2024osprey} & 100K & 503K & 724K & \cmark & \cmark/\xmark & &\cmark/\cmark & & \cmark & - \\
GranD \cite{rasheed2024glamm} & 11M & 810M & 7.5M & \cmark & \cmark/\cmark &
&\cmark/\cmark & & \xmark & -\\
MUSE \cite{ren2024pixellm} & 246K & 910K & 246K & \cmark & \cmark/\xmark &
&\cmark/\cmark & & \cmark & - \\
\rowcolor{lightgray}
\multicolumn{11}{c}{\emph{Multimodal Reasoning Benchmark}} \\
ScienceQA \cite{lu2022learn} & 5.6K & - & 5.6K & \cmark & \cmark/\cmark &
&\xmark/\xmark & & \cmark & 2.5 \\
MMMU \cite{yue2024mmmu} & 11.5K  & - & 11.5K & \cmark & \cmark/\xmark &
&\xmark/\xmark & & \cmark & 1.0 \\
$\text{M}^3\text{COT}$ \cite{chen2024m} & 11K & - & 11K & \cmark & \cmark/\xmark & &\xmark/\xmark & & \cmark & 10.9 \\ \midrule
\rowcolor{lightorange}
PRIST (Ours) & 2.8K & 8.3K & 8.3K & \cmark & \cmark/\cmark & &\cmark/\cmark & & \cmark & 4.0 \\
\bottomrule
\end{tabularx}
\caption{Comparison of existing segmentation and multimodal reasoning benchmarks. \textbf{\#X}: the size of X,
\textbf{Img.:} Image; \textbf{Reg.:} Segmentation Regions; \textbf{Samp.:} Conversational Samples; \textbf{Step:} Steps in the reasoning chain, averaged over all samples.}
\label{tab:compare_benchmark}
\end{table*}
\subsection{Data Generation Process} \label{sec:appendixA.2}
\begin{figure*}[htbp]
    \centering
    \captionsetup{skip=0pt}  
    \begin{subfigure}[b]{\textwidth}
        \centering
        \captionsetup{skip=-2pt}  
        \includegraphics[width=\textwidth]{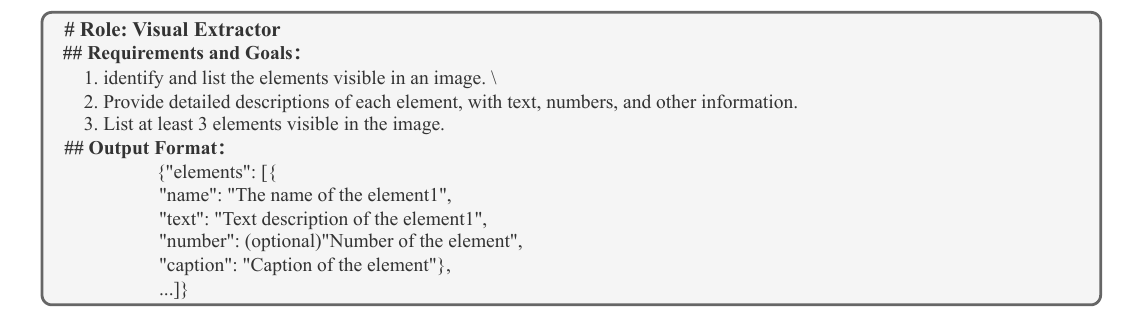}
        \caption{Illustration of the visual elements extraction (step-1).}
        \label{fig:sub1}
    \end{subfigure}
    \begin{subfigure}[b]{\textwidth}
        \centering
        \captionsetup{skip=-8pt}
        \includegraphics[width=\textwidth]{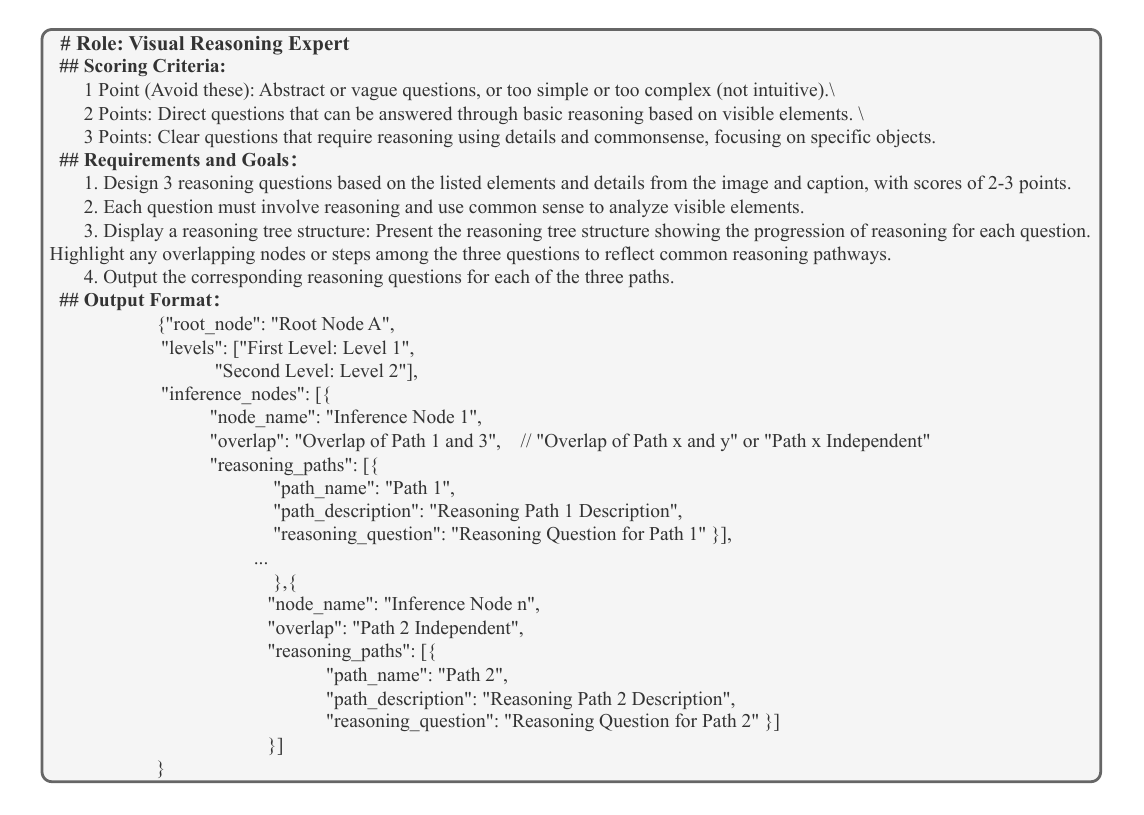}
        \caption{Illustration of the reasoning tree construction (step-2).}
        \label{fig:sub2}
    \end{subfigure}
    \begin{subfigure}[b]{\textwidth}
        \centering
        \captionsetup{skip=-5pt}
        \includegraphics[width=\textwidth]{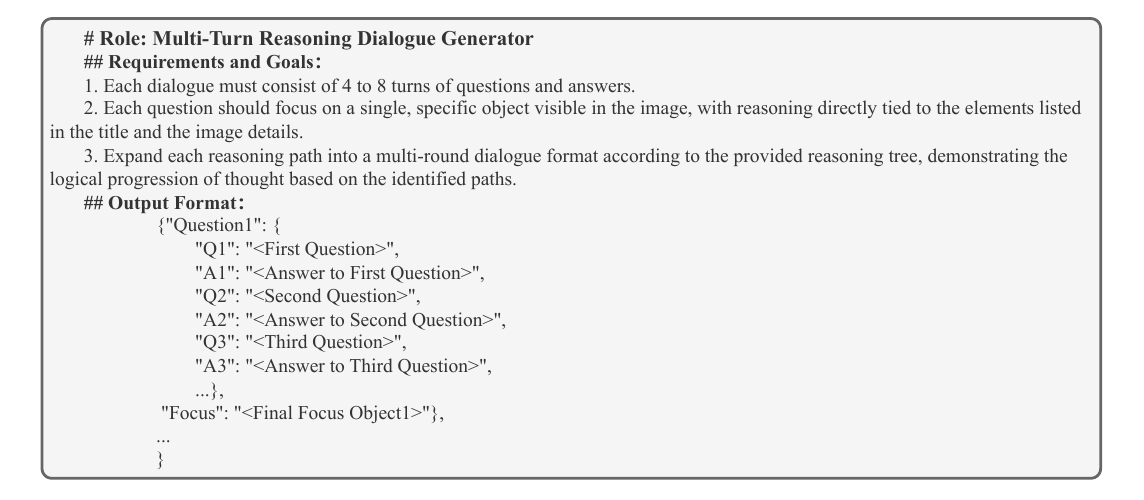}
        \caption{Illustration of the multi-turn dialogue generation (step-3).}
        \label{fig:sub3}
    \end{subfigure}
    \caption{The prompts and output formats of our dataset annotation pipeline.}
    \label{fig:pipeline_prompt}
\end{figure*}

We design a fully automated dataset annotation pipeline, leveraging multiple hierarchical levels in the visual domain to construct the PRIST dataset. The pipeline, entirely based on GPT-4o (\textit{gpt-4o-2024-08-06}), incorporates CoT into a feedback loop to generate relevant multi-turn reasoning dialogues for various images. Each dialogue’s final query is a segmentation-related instruction (e.g., \textit{"Please segment the core objects according to the above dialogue"}), making PRIST suitable for pixel-level segmentation tasks and extending to general VQA (when ignoring the final instruction), providing a versatile foundation for multimodal dialogue research. The data is used to fine-tune MLLMs. We will release the PRIST dataset and the implementation of its automated annotation pipeline to support further research. For detailed information on the implementation of LLM prompts and output formats at three levels, refer to Figure \ref{fig:pipeline_prompt}.

\subsection{Dataset Statistics and Analysis} \label{dataappendix}
\begin{figure}
    \centering
    \captionsetup{skip=1.5pt}
    \includegraphics[width=1\linewidth]{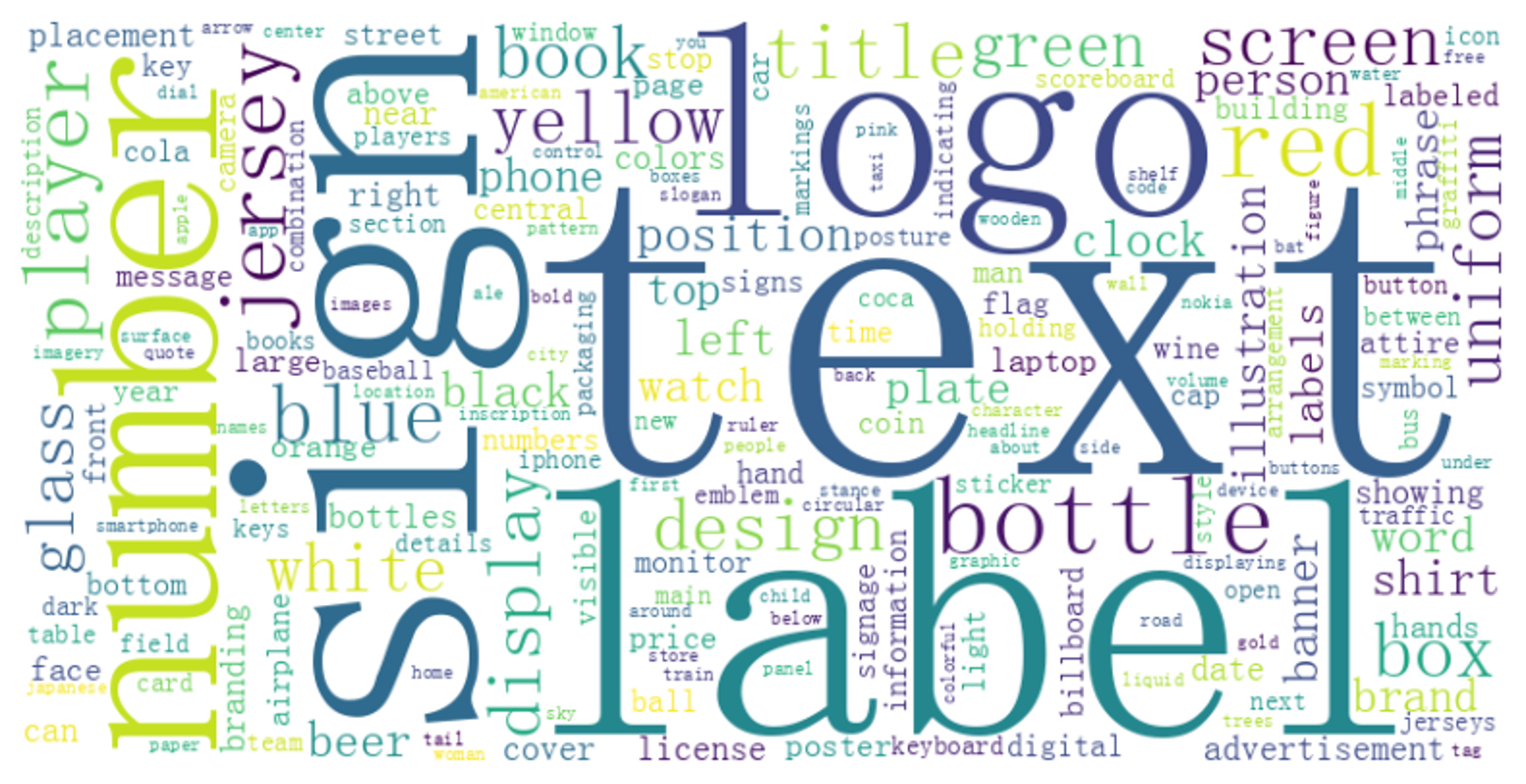}
    \caption{Wordcloud of the 200 popular focus-related words in PRIST.}
    \label{fig:wordcloud}
\end{figure}

As shown in Table \ref{tab:compare_benchmark}, PRIST is compared with existing segmentation and multimodal reasoning benchmarks. PRIST combines the double advantages of pixel-level segmentation and multi-turn reasoning. In comparison with segmentation benchmarks \cite{yu2016modeling,krishna2017visual,lai2023lisa,rasheed2024glamm}, the focus is on the granularity and richness of the segmentation targets. Figure \ref{fig:wordcloud} illustrates a visual word cloud of the 200 most popular focus-related terms extracted from PRIST. The prominence of each word in the cloud represents its frequency of occurrence in the dataset.

Compared to existing multimodal reasoning benchmarks \cite{lu2022learn,zhang2023multimodal,chen2024m}, PRIST enables a more in-depth and detailed reasoning process for each segmented object. The average text length per sample is 477 tokens, significantly higher than the 294 tokens in $\text{M}^3\text{CoT}$ \cite{chen2024m}.  Each conversation contains an average of 4 turns, with a maximum of 7, surpassing ScienceQA's average of 2.5 \cite{lu2022learn}. PRIST better simulates real-world interactions and human thought processes while ensuring more coherent and detailed reasoning chains for pixel-level reasoning segmentation, presenting a new challenge for the RS field.
\section{Experiment Setups} \label{setups}
\subsection{Implementation Details} \label{trainsetting}
Our experiments are all conducted on 2 NVIDIA A6000 (48G) GPUs. Aligned with region-level segmentation models \cite{rasheed2024glamm,zhang2024omg}, we adopt LLaVAv1.6-7B \footnote{\url{https://huggingface.co/llava-hf/llava-v1.6-vicuna-7b-hf}} as the backbone $\mathcal{F}$ and ViT-H SAM \footnote{\url{https://github.com/facebookresearch/segment-anything}} to instantiate pixel-level encoder $\mathcal{V}_{\text{pixel}}$ and mask decoder $\mathcal{D}_m$. The visual-language projection layer is implemented through 2-layer MLP and GELU activation following LLaVA-v1.6. The implementation is carried out in PyTorch, with Deepspeed Zero-2 optimization applied during two-stage training, and LoRA fine-tuning with $r=8$ for the LLM. Different experimental setups are adopted at each training stage to facilitate model convergence. Detailed training configurations are provided in Table \ref{tab:config}.
\begin{table}[h!]
    \centering
    \setlength{\tabcolsep}{3.8pt} 
    \begin{tabularx}{0.49\textwidth}{l|cc}
    \toprule
       Config  & Stage 1 & Stage 2 \\ \midrule
       Optimizer & \multicolumn{2}{c}{AdamW} \\
       Optimizer momentum & \multicolumn{2}{c}{$\beta_1=0.9,\beta_2=0.95$}\\
       Learning rate schedule & \multicolumn{2}{c}{WarmipDecayLR} \\
       Warmup iterations & \multicolumn{2}{c}{100} \\
       Weight decay & \multicolumn{2}{c}{0} \\
       Gradient accumulation steps & \multicolumn{2}{c}{10} \\\midrule
       Learning rate & 3e-4 & 1e-5 \\
       Batch size & 16 & 32 \\
       Training steps & 50k & 20k \\ 
       \bottomrule
    \end{tabularx}
    \caption{The training settings of MIRAS.}
    \label{tab:config}
    \vspace{-15pt}
\end{table}
\subsection{Two-Stage Training} \label{two-stage}
We employ a two-stage training process to achieve efficient pixel-level reasoning segmentation. The specific details of each stage are as follows:
\vspace{-3pt}
\paragraph{Stage 1: Mask-Text Alignment Pre-training}
The objective of this stage is to align mask-based region features with language embeddings. We collect mask-text pairs from various publicly available object-level datasets, including COCO \cite{chen2015microsoft}, Ade20k \cite{zhou2019semantic}, and Mapillary \cite{neuhold2017mapillary}, as well as region-level datasets like the RefCOCO series \cite{yu2016modeling,kazemzadeh-etal-2014-referitgame} and PACO \cite{ramanathan2023paco}. Additionally, we incorporate VQA and caption data, such as LLaVA-Instruct-80k \cite{liu2024improved}, COCO Caption\cite{chen2015microsoft}. We mix these data in a 9:6:2:2 ratio and convert them into an instruction-following format for training, thereby enhancing its perceptual and conversational abilities.

\vspace{-3pt}
\paragraph{Stage 2: Pixel-level Segmentation Fine-tuning}
At this stage, we maintain fixed model weights while fine-tuning our PRIST dataset to enhance fine-grained segmentation and reasoning capabilities. Subsequently, we integrate VQA data \cite{liu2024improved} to enable MIRAS to follow user instructions (i.e., PRIST combined with VQA at a 4:1 ratio), enhancing its ability to handle complex pixel-level segmentation with precision.
\begin{tcolorbox}[title = {Segmentation Prompt}] 
\textbf{Output Format:} 

<box>(x1, y1), (x2, y2)</box> 
\tcblower
Please box out the position of {focus} and output the detection box in <box>(x1, y1), (x2, y2)</box> format, with coordinates representing the top-left and bottom-right corners of the detected area. If no focus is detected, return "no detected", and ensure a blank space follows all outputs.
\end{tcolorbox}
\subsection{General MLLMs Baselines} \label{generalmllm—setup}
To evaluate the advanced general MLLMs on our task, we guide them with a unified segmentation prompt. Since these models do not directly generate masks, we first extract the bounding box coordinates and convert them into masks using a consistent function. The generated masks are then compared with ground truth masks. For fairness, we exclude the last round of dialogue in the final evaluation, as the models were not trained with identical segmentation instructions.

\subsection{Evaluation} \label{sec:appendixB.2}
\subsubsection{Pixel-level Segmentation Metrics} \label{segmetrics}
We adopt traditional methods for calculating precision and recall, applying them to the pixel-wise task. Specifically, each pixel is treated as a binary classification (\text{i.e.,} 0,1) to quantify the model's ability in pixel-level segmentation.

\noindent\textbf{\textit{Pixel-wise Precision}} measures the proportion of true positive samples among the pixels predicted as positive, reflecting the model's prediction accuracy. Its increase indicates the model can predict the pixels of the target region with higher confidence, thereby capturing the target more precisely.

\begin{small}
\begin{equation*}
Precision = \frac{TP}{TP + FP}
\end{equation*}
\end{small}%
where $TP$ represents pixels correctly predicted as the target, while $FP$ represents non-target pixels incorrectly predicted as the target.

\noindent\textbf{\textit{Pixel-wise Recall}} measures the proportion of actual positive samples that are correctly predicted as positive, reflecting the model's coverage of the target area. Its decrease indicates that the model is more strictly segmenting according to the target boundaries, thereby avoiding overgeneralization.

\begin{small}
\begin{equation*}
Recall = \frac{TP}{TP + FN}
\end{equation*}
\end{small}%
where $TP$ represents the pixels correctly predicted as the target, while $FN$ represents the non-target pixels that are incorrectly predicted as the target.
\subsubsection{LLM-based Metrics} \label{llm_metrics}
\begin{figure*}
    \centering
    \captionsetup{skip=1.5pt}
    \includegraphics[width=1\linewidth]{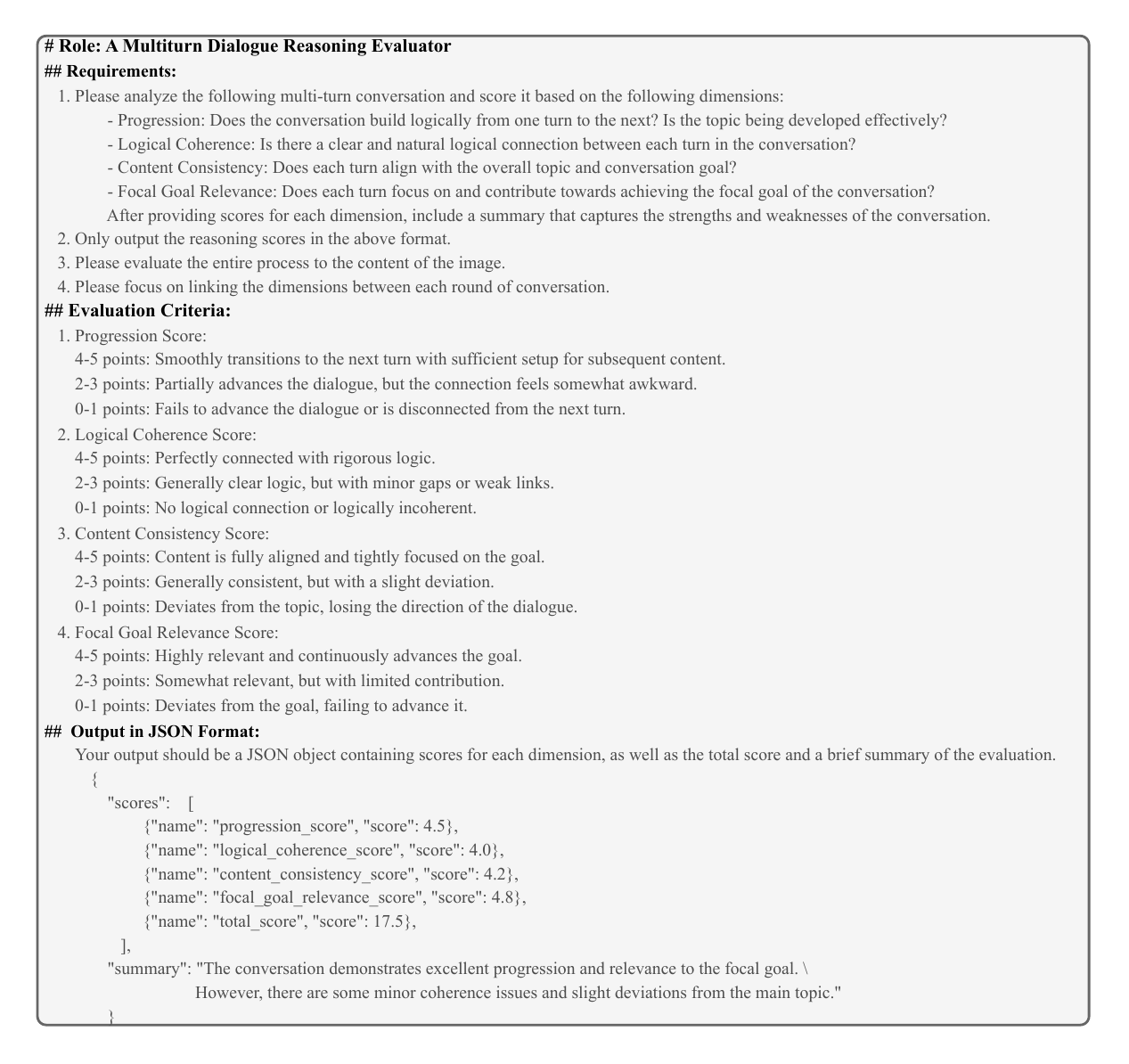}
    \caption{Illustration of the LLM-based evaluation of reasoning quality.}
    \label{fig:llmeval}
\end{figure*}

Traditional metrics (e.g., BLEU-n, METEOR) often fail to capture deep semantics and logical coherence in dialogue. To address this, we employ GPT-4 as a scoring tool to evaluate the model's reasoning quality using four key metrics: \textbf{\textit{1) Progressiveness (PR):}} Whether the current turn effectively sets up and guides the next. \textbf{\textit{2) Logical Coherence (LC):}} Whether there is a smooth logical connection between the current and subsequent turns. \textbf{\textit{3) Content Consistency (CC):}} Whether the dialogue revolves around the overall topic. \textbf{\textit{4) Target Relevance (TR):}} Whether the dialogue stays focused on the target. We carefully designed prompts for each metric to ensure fairness and consistency. To ensure fairness and consistency, we carefully designed prompts for each metric. Ratings are assigned on a 5-point scale divided into three intervals (0-1, 2-3, 4-5), with clearly defined criteria for each range. The final score is calculated as the average across all four metrics, minimizing randomness and bias. Higher scores indicate better performance, reflecting overall improvement across evaluated aspects. GPT-4 strictly adheres to these guidelines, evaluating each dimension step by step. 
Notably, when using an LLM for scoring, to eliminate randomness, we repeat the scoring for each metric 5 times and calculate the average to determine the final score. Detailed evaluation prompts are shown in Figure \ref{fig:llmeval}. 

\subsubsection{Human Response}\label{human} We randomly select 100 multi-turn conversations from the PRIST test set and invite three experts in computer science to respond masking the original answers. They all possess solid analytical skills and a deep understanding of model reasoning tasks. To ensure fairness in the evaluation, we also used the LLM-based evaluation criteria to score the three responses (as shown in Table \ref{tab:humanscore}). Finally, the average score from the three experts is used as the human evaluation value for each metric.

\begin{table}[h!]
    \centering
    \setlength{\tabcolsep}{6pt} 
    \begin{tabularx}{0.4\textwidth}{lcccc}
    \toprule
     & \textbf{PR} & \textbf{LC} & \textbf{CC} & \textbf{TR} \\ \midrule
    Expert 1 & 4.06 & 4.10 & 4.38 & 4.29 \\
    Expert 2 & 3.98 & 4.02 & 4.21 & 4.31 \\
    Expert 3 & 4.05 & 4.00 & 4.19 & 4.24 \\ \midrule
    \textbf{Human} & 4.03 & 4.04 & 4.26 & 4.28 \\
    \bottomrule
    \end{tabularx}
    \caption{The reasoning scores of three human experts based on LLM-based metrics.}
    \label{tab:humanscore}
\end{table}

Additionally, we leverage GPT-4 as a judge to design an adversarial evaluation framework, using \textbf{Win Rate (\%)} to assess reasoning quality in multi-turn dialogues. Human responses serve as the standard for model output comparison, the model is considered to achieve a win when its score surpasses the human response score.

\begin{figure}
    \centering
    \captionsetup{skip=1.5pt}
    \includegraphics[width=1\linewidth]{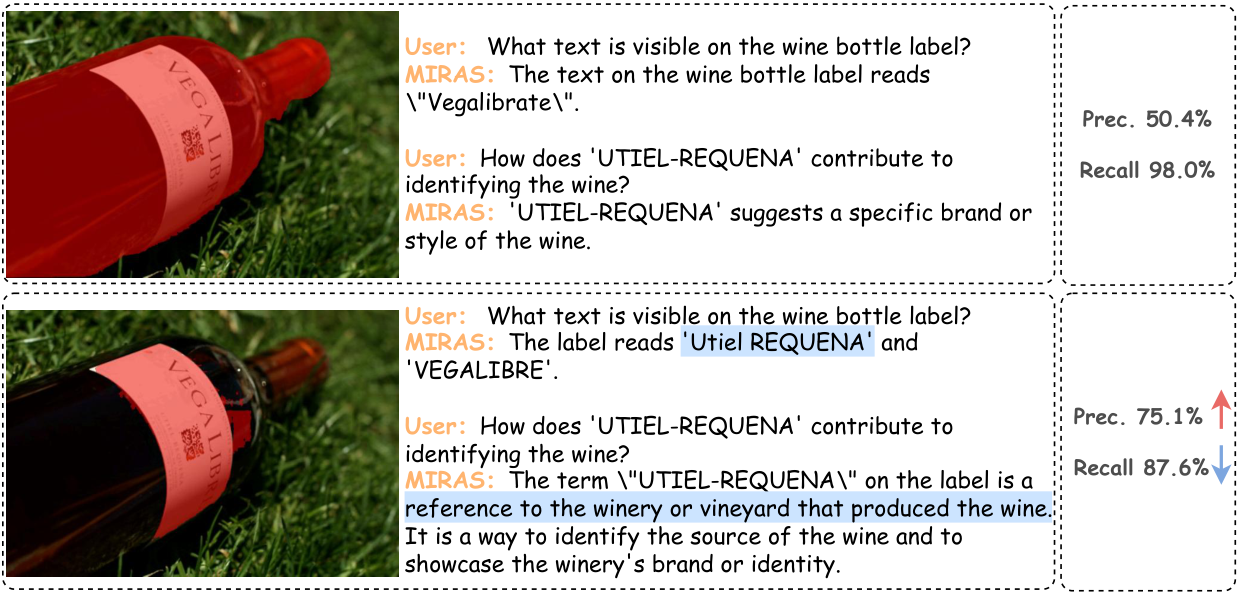}
    \caption{Qualitative results of the performance of MIRAS on Pixel-level RS. \textbf{Top:} Stage-1 segments \textit{\textcolor{blue}{the entire bottle}}. vs. \textbf{Bottom:} Stage-2 only segments \textit{\textcolor{red}{the label}} and provides more accurate and comprehensive interpretations of \textit{the “UTIEL-REQUENA” label}.}
    \label{fig:qualitative result}
    \vspace{-15pt}
\end{figure}

\section{Case Study} \label{casestudies}

\begin{figure*}[htbp]
    \centering
    \captionsetup{skip=-5pt}  
    \begin{subfigure}[b]{\textwidth}
        \centering
        \captionsetup{skip=-2pt}  
        \includegraphics[width=\textwidth]{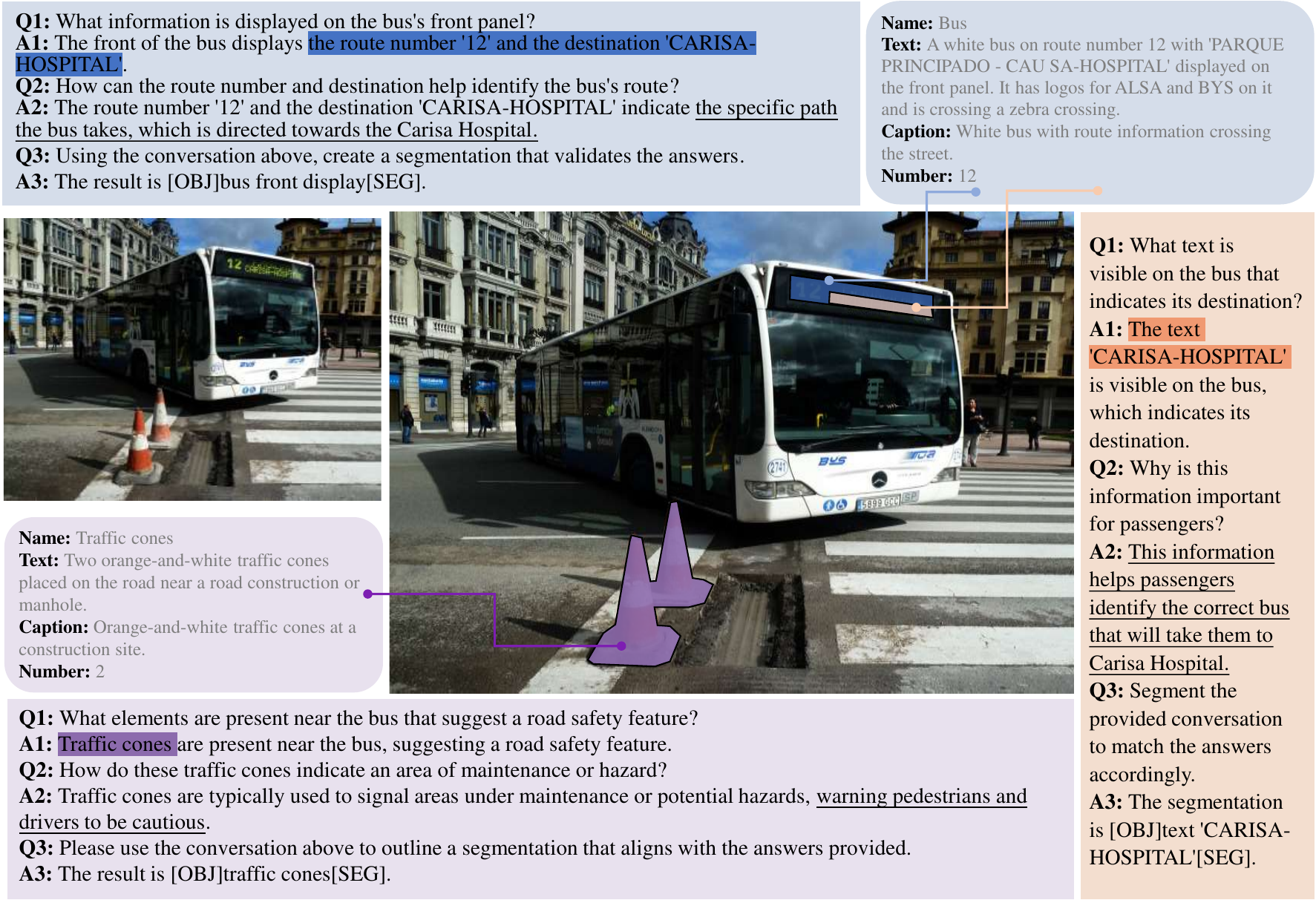}
        \label{fig:exam_sub1}
    \end{subfigure}
    \begin{subfigure}[b]{\textwidth}
        \centering
        \captionsetup{skip=-8pt}
        \includegraphics[width=\textwidth]{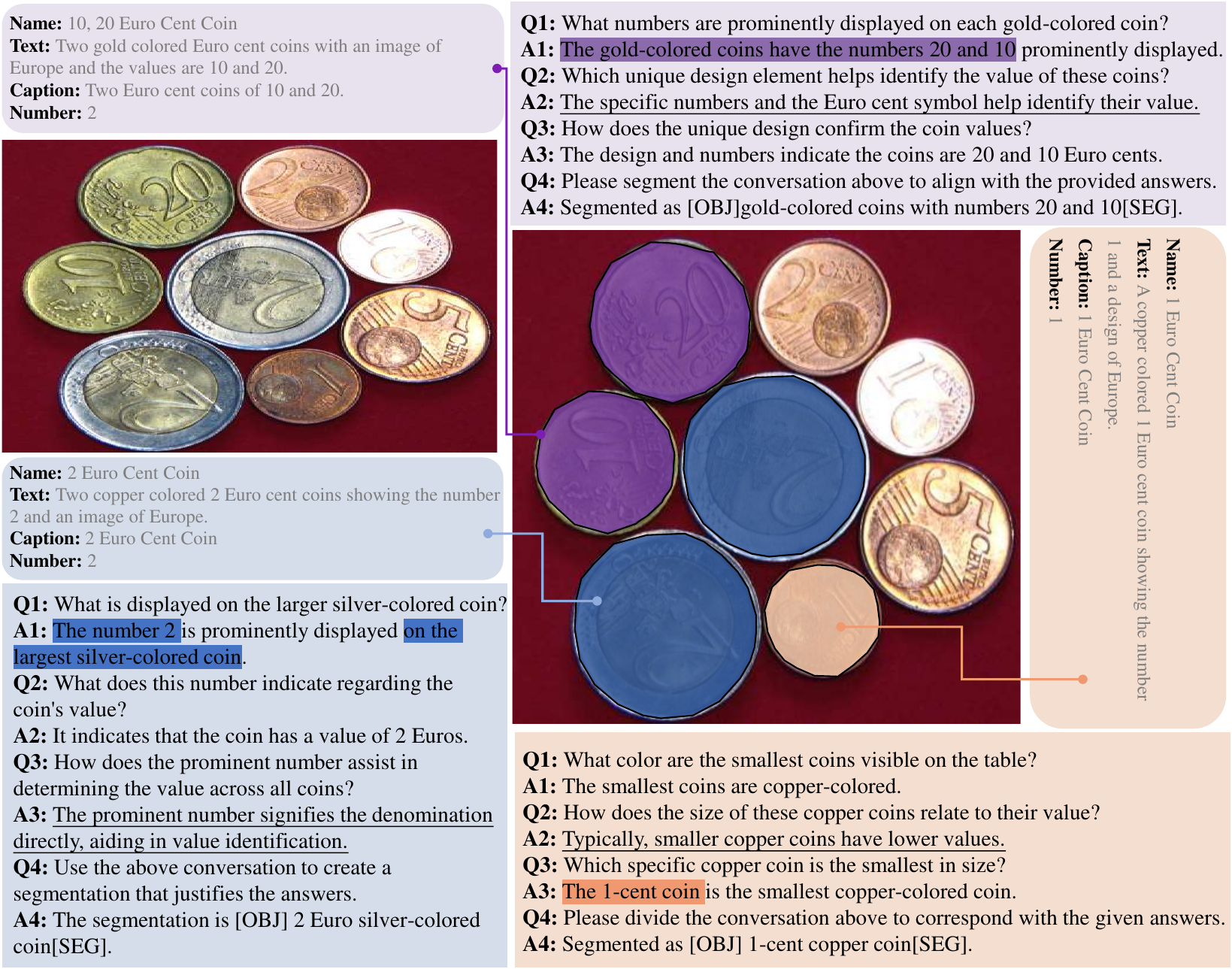}
        \label{fig:exam_sub2}
    \end{subfigure}
    \caption{\textbf{The samples of PRIST.} The figure displays two examples from the PRIST dataset, created through the automated annotation pipeline. It offers a range of semantic labels and attributes for the identified objects, as well as contextualized multi-turn conversations.}
    \label{fig:prist_example}
\end{figure*}

To reflect the high quality of our constructed dataset, we present examples of PRIST test data in Figure \ref{fig:prist_example}, which effectively combine coherent and logically structured multi-turn conversations with fine-grained segmentation. We fine-tuned MIRAS on the PRIST dataset to enhance its understanding and localization of image details. As shown in Figure \ref{fig:qualitative result}, the fine-tuned MIRAS (\textit{Stage-2}) segments only \textit{the label}, replacing \textit{the entire bottle} in Stage-1, demonstrating stronger segmentation specificity (with a significant improvement in precision) while effectively mitigating the issue of overgeneralization (with a slight decrease in recall). This result also highlights the potential of PRIST in improving pixel-level reasoning capabilities.

\begin{figure*}[t]
    \centering
    \includegraphics[width=1\linewidth]{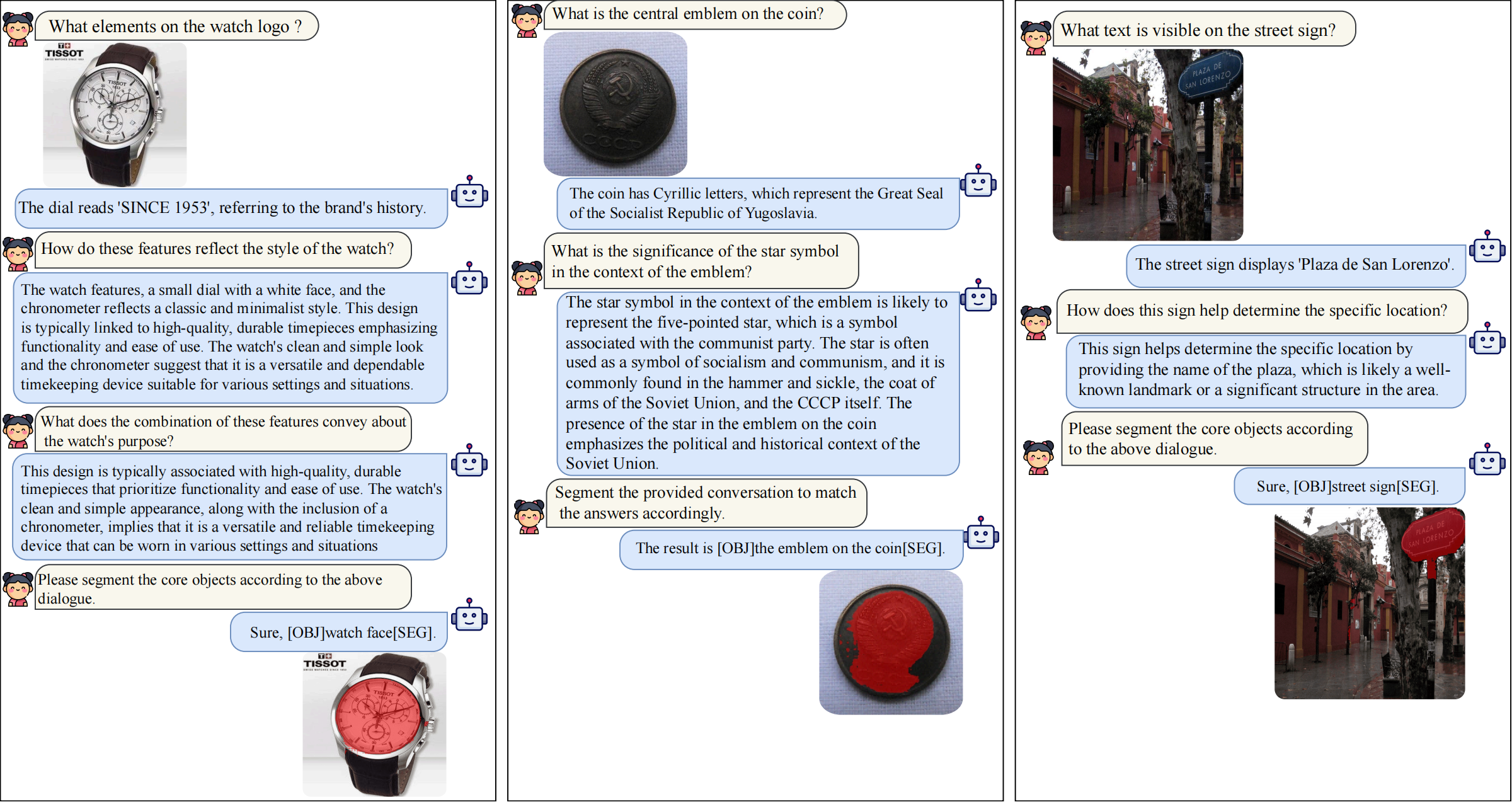}
    \caption{Pixel-level segmentation and multi-turn conversational interactions facilitated by MIRAS. \textbf{Left:} The model segments only the watch face instead of the entire watch, and identifies the three sub-dials and the "SINCE 1953" logo. \textbf{Center:} The model segments only the emblem on the coin instead of the whole coin, and infers that it is related to Soviet communism. \textbf{Right:} The model segments only the road sign, and accurately recognizes the text "Plaza de San Lorenzo".}
    \label{fig:casestudyall}
\end{figure*}

To more intuitively demonstrate the advantages of our model in the pixel-level reasoning segmentation task, we also present several cases of user interactions with the chatbot. As shown in Figure \ref{fig:casestudyall}, these cases illustrate MIRAS's coherence and consistency in multi-turn interactions and showcase its ability to achieve more precise fine-grained object segmentation.

\end{document}